\documentclass{article}

\usepackage{amsmath,amsfonts,bm}

\def\eqref#1{equation~\ref{#1}}
\def\1{\bm{1}}

\DeclareMathAlphabet{\mathsfit}{\encodingdefault}{\sfdefault}{m}{sl}
\SetMathAlphabet{\mathsfit}{bold}{\encodingdefault}{\sfdefault}{bx}{n}

\usepackage{xcolor}

\renewcommand{\paragraph}[1]{\vspace{0.2cm}\noindent\textbf{#1}}

\definecolor{midnightgreen}{rgb}{0.0, 0.29, 0.33}
\newcommand{\cx}[1]{\textcolor{midnightgreen}{\bf\small [#1 --cx]}}
\renewcommand{\cx}[1]{}

\usepackage{hyperref}
\definecolor{codegreen}{rgb}{0,0.6,0}
\definecolor{prevcitegreen}{rgb}{0.0, 0.42, 0.24} %
\definecolor{citegreen}{rgb}{0.0, 0.42, 0.24} % Color reference: https://openreview.net/pdf?id=DJSZGGZYVi
\definecolor{codegray}{rgb}{0.5,0.5,0.5}
\definecolor{codepurple}{rgb}{0.58,0,0.82}
\definecolor{backcolour}{rgb}{0.95,0.95,0.92}

\definecolor{midnightblue}{rgb}{0.11, 0.11, 0.6} % 
\definecolor{rebuttal}{rgb}{0.0, 0.0, 1.0}
\definecolor{rebuttal2}{rgb}{0.0, 0.0, 0.0}
\hypersetup{
    colorlinks=true, % Enables colored links
    linkcolor=midnightblue,  % Color of internal links (e.g., Table of Contents)
    citecolor=midnightblue,   % Color of citations (works with \citep and \citet)
}

\usepackage{microtype}
\usepackage{graphicx}
\usepackage{subcaption}
\usepackage{booktabs} % for professional tables
\usepackage{tabularx}
\usepackage{wrapfig}
\usepackage{colortbl}
\usepackage{adjustbox}
\usepackage{multirow}
\usepackage{makecell}
\usepackage[most]{tcolorbox}
\usepackage{url}
\usepackage{etoc}

\usepackage{hyperref}

\usepackage[accepted]{icml2026}

\usepackage{amsmath}
\usepackage{amssymb}
\usepackage{mathtools}
\usepackage{amsthm}

\usepackage[capitalize,noabbrev]{cleveref}

\theoremstyle{plain}

\theoremstyle{definition}

\theoremstyle{remark}

\usepackage[textsize=tiny]{todonotes}

\icmltitlerunning{RePro: Training Language Models to Faithfully Recycle the Web for Pretraining}

\begin{document}

\twocolumn[
  \icmltitle{RePro: Training Language Models to Faithfully Recycle the Web for Pretraining}



  \icmlsetsymbol{equal}{*}

  \begin{icmlauthorlist}
    \icmlauthor{Zichun Yu}{yyy}
    \icmlauthor{Chenyan Xiong}{yyy,comp}
    % \icmlauthor{Firstname1 Lastname1}{equal,yyy}
    % \icmlauthor{Firstname2 Lastname2}{equal,yyy,comp}
    % \icmlauthor{Firstname3 Lastname3}{comp}
  \end{icmlauthorlist}

  \icmlaffiliation{yyy}{Language Technologies Institute, Carnegie Mellon University}
  \icmlaffiliation{comp}{Xlue}
  %   \icmlaffiliation{sch}{School of ZZZ, Institute of WWW, Location, Country}

  \icmlcorrespondingauthor{Zichun Yu}{zichunyu@andrew.cmu.edu}
  \icmlcorrespondingauthor{Chenyan Xiong}{cx@andrew.cmu.edu}

  \icmlkeywords{Large Language Model, Pretraining, Synthetic Data, Machine Learning, ICML}

  \vskip 0.3in
]



\printAffiliationsAndNotice{}  % no special notice (required even if empty)

\begin{abstract}
    % High-quality pretraining data is the fossil fuel of large language models (LLMs), yet its reserves are running low for frontier models. 
    % Existing rephrasing-based methods are able to recycle moderate- to low-quality web data, but they typically rely on prompting large models and offer little control over faithfully preserving the original content. 
    High-quality data is a cornerstone of large language model (LLM) pretraining, yet its growth has not kept pace with the needs of frontier models.
    In this paper, we introduce \textsc{RePro}, a novel web recycling method that trains a relatively small LM with reinforcement learning to generate effective and faithful rephrasings of pretraining data. Specifically, we design one \textit{quality} reward and three \textit{faithfulness} rewards, optimizing the LM rephraser to convert organic data into high-quality rephrasings while maintaining its core semantics and structure. 
    In our experiment,
    we train a rephraser as small as 1B parameters to recycle 72B tokens sampled from DCLM-RefinedWeb. 
    % and we apply DCLM-fastText classifier on both organic and recycled data to select the high-quality portion.  
    Pretraining results on 400M, 1.4B, and 2.8B models demonstrate that \textsc{RePro} delivers 3.7\%-14.5\% relative accuracy gains over organic-only baseline on 22 downstream tasks, doubling the performance gains achieved by the state-of-the-art web recycling method that prompts a 70B rephraser.
    % \textsc{RePro} also outperforms ReWire, the state-of-the-art web recycling method that prompts a 70B rephraser, as well as the organic baseline with a 4$\times$ larger data pool. Experiments with different amounts of recycled data highlight that \textsc{RePro} improves organic data efficiency by 2-3$\times$.
    % triples the usable supply of high-quality training data. 
    Individual and distributional analyses validate that \textsc{RePro} preserves more critical information and faithfully reflects the characteristics of organic data compared to prompting-based methods. Together, these results show that \textsc{RePro} provides an efficient and controllable path to effectively recycle organic data for pretraining.
    % ``fossil fuel'' of LLM pretraining.
    Our code is available at \url{https://github.com/cxcscmu/RePro}. 
    % We will open-source our rephraser and recycled data.
\end{abstract}
% The internet, often described as the “fossil fuel” of AI—offers only a finite reservoir of training data.
\section{Introduction}

% The era of unlimited web-crawled content (organic data) for pretraining is nearing its end, 
% causing the scaling of large language models (LLMs) to plateau~\citep{villalobos2024will,maini2025beyondweb}.
% Standard data cleaning pipelines often filter out more than 90\% of organic data~\citep{weber2024redpajama,li2024datacomp},
% leading to a looming ``data wall'' for LLM pretraining.

% \begin{wrapfigure}
%     {r}{0.6\textwidth}
\begin{figure}
    \centering
    % \vspace{-16pt}
    \begin{subfigure}[t]
        {0.24\textwidth}
        \centering
        \raisebox{0pt}{\includegraphics[width=1.0\linewidth]{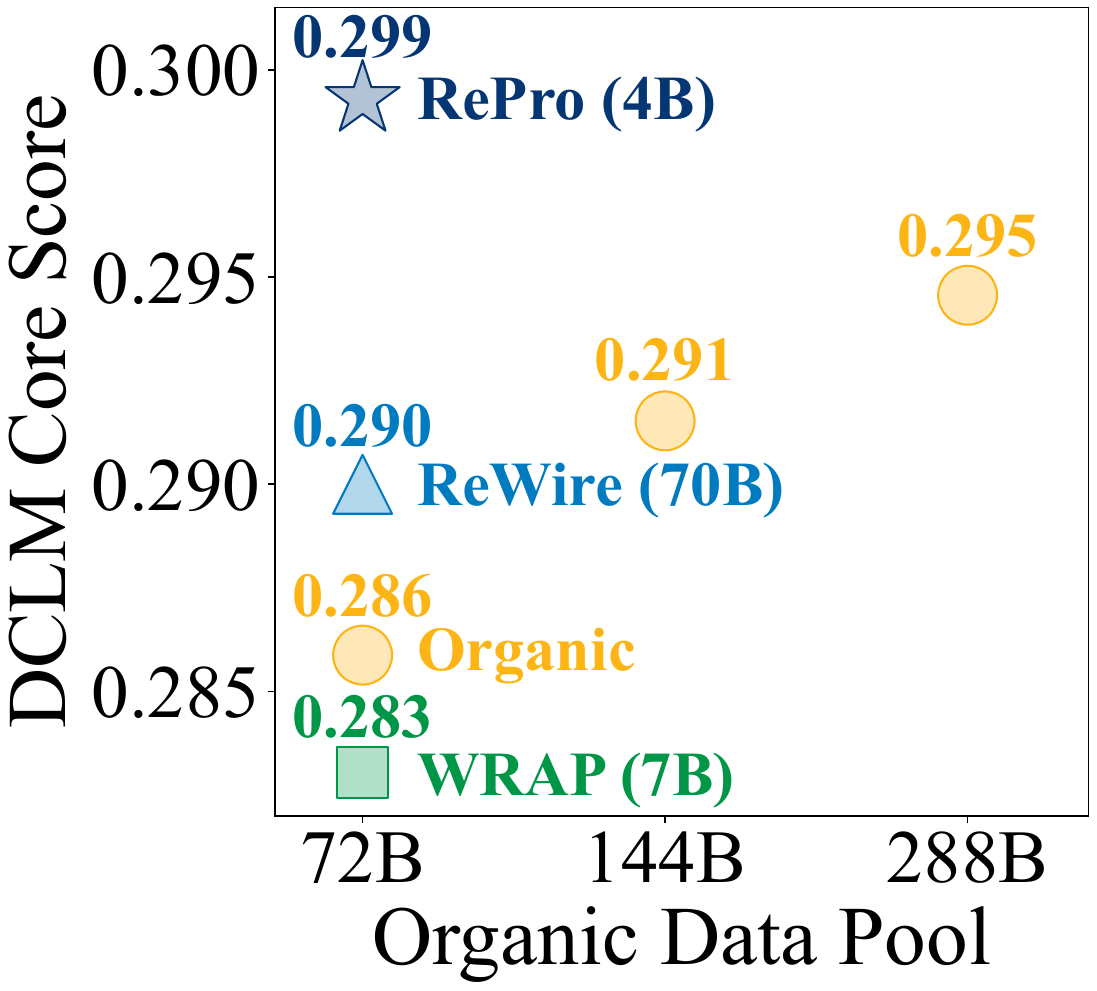}}
        % \vspace{-11pt}
        \caption{Downstream performance}
        \label{fig:overall-cmp}
    \end{subfigure}
    ~
    \begin{subfigure}[t]
        {0.218\textwidth}
        \centering
        \includegraphics[width=1.0\linewidth]{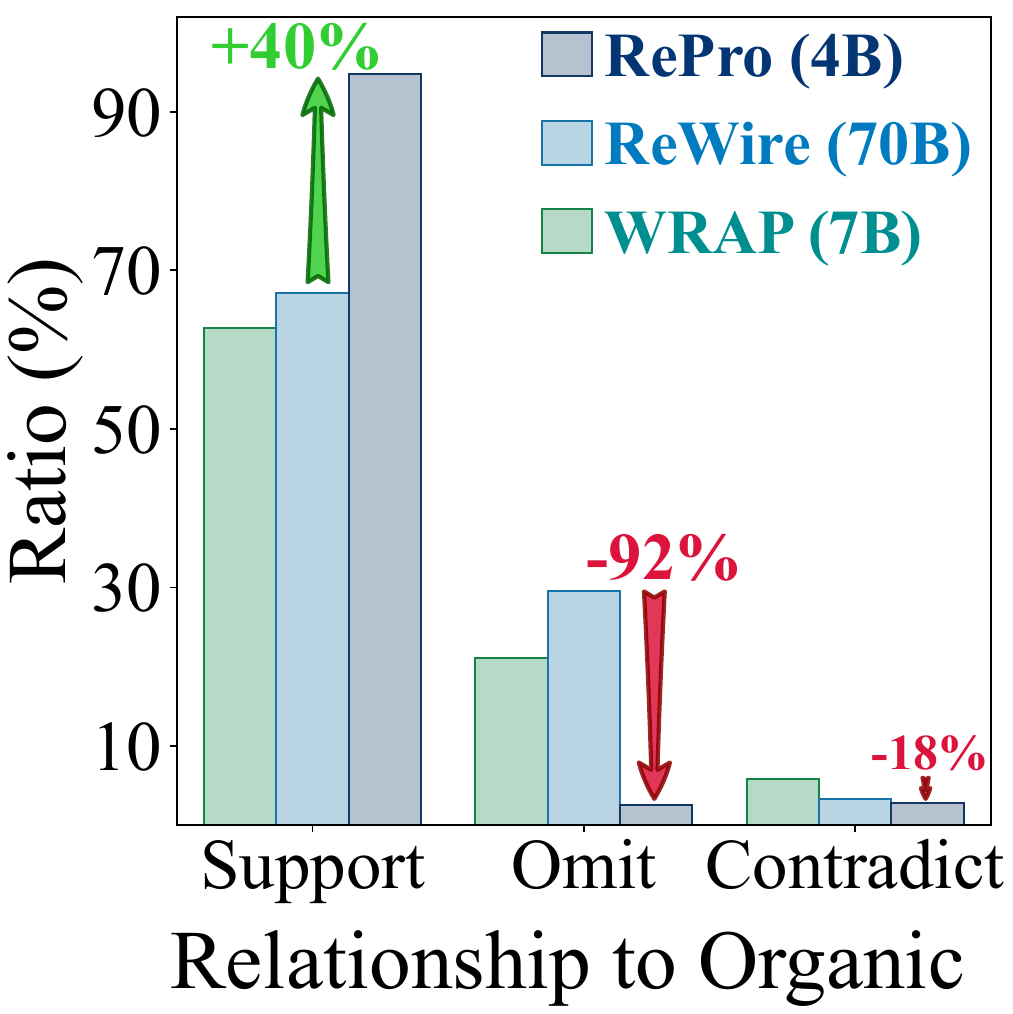}
        % \vspace{-0.35cm}
        \caption{Key point recall}
        \label{fig:keypoint-recall}
    \end{subfigure}
    \vspace{-0.1cm}
    \caption{(a) Pretraining performance of 1.4B model and (b) ratio of key points in organic data that are supported / omitted / contradicted in the recycled version across three web recycling methods, computed on 30{,}000 randomly sampled instances. The rephraser size in denoted in parentheses.}
    \vspace{-0.6cm}
    \label{fig:clustering}
\end{figure}
% \end{wrapfigure}

The continued scaling of large language models (LLMs) is threatened by the diminishing supply of high-quality pretraining data~\citep{villalobos2024will,maini2025beyondweb}.
While the web provides vast amounts of content, referred to as organic data, 
standard data pipelines often filter out much of it as ``low-quality'' to ensure pretraining quality~\citep{weber2024redpajama,li2024datacomp}.
% However, the identified low-quality data is not necessarily useless, e.g., 
% a poorly formatted article may still contain valuable knowledge.
% This aggressive filtering amplifies the data scarcity problem and thus leads to a looming ``data wall'' that 
The shortage of high-quality data thus leads to a looming ``data wall'' that impedes further progress in LLM pretraining~\citep{nguyen2025rewire}.
% The shortage of high-quality data—the fossil fuel of pretraining—
% impedes further progress in scaling frontier LLMs~\citep{maini2025beyondweb}.

% Rephrasing web data by LLMs is an effective workaround to 
% recycle low-quality data and thus increase the quantity of usable pretraining data~\citep{nguyen2025rewire}. 
% Previous rephrasing methods by prompting LLMs (e.g., Llama-3.3-70B-Instruct~\citep{dubey2024llama}) 
% show promising results, which can be comparable or even more effective than having access to 2$\times$ organic data~\citep{maini2024rephrasing,nguyen2025rewire}.
% However, there are two main limitations in these methods. 
% First, the huge cost of rephrasing with large models is a great concern, 
% which often demands orders of magnitude more compute than pretraining itself~\citep{nguyen2025rewire}.
% Second, prompting-based rephrasing lacks nuanced control of the behavior of the rephraser, 
% as a generally stronger model may not be the better rephraser to 
% generate high-quality and diverse data for pretraining~\citep{maini2025beyondweb}.
% Hence, it is crucial to develop a cost-efficient and controllable rephrasing method to 
% better recycle web data for LLM pretraining.

To address the scarcity of high-quality data, one promising path is to ``\textit{recycle}'' the low-quality web data by rephrasing, thereby increasing the amount of usable pretraining data~\citep{nguyen2025rewire}.
Previous approaches by prompting LLMs (e.g., Llama-3.3-70B-Instruct~\citep{dubey2024llama}) 
have shown promising results, matching or even surpassing the performance of doubling organic data~\citep{maini2024rephrasing,nguyen2025rewire}.
However, these methods face two major limitations.
First, the computational cost of rephrasing with large models is prohibitively high; 
% often demanding orders of magnitude more compute than pretraining itself~\citep{nguyen2025rewire}.
second, 
% prompt-based rephrasing offers limited control over the behavior of the rephraser: 
% a strong general-purpose model is not necessarily a fit one for generating high-quality and diverse pretraining data~\citep{maini2025beyondweb}.
prompts alone may not faithfully preserve the semantics and structure of organic data~\citep{bi2025refinex}, which is critical for the reliability and richness of the pretraining corpus.
These challenges underscore the need for a cost-efficient and faithful web recycling method for LLM pretraining.

In this paper, we introduce \textsc{RePro}, an efficient and faithful web recycling method to effectively harness organic pretraining data. Our approach leverages reinforcement learning (RL) to train a relatively small rephraser using two categories of rewards: \textit{quality} and \textit{faithfulness}. Specifically, we choose DataMan score~\citep{peng2025dataman}, a holistic quality assessment metric, as the quality reward to incentivize high-quality rephrasings. For faithfulness rewards, we employ BERTScore~\citep{zhang2020bertscore}, structure alignment, and length preservation to better approximate organic data. Together, these rewards guide the rephraser to generate high-quality data while faithfully preserving the core properties of organic data.

% We empirically verify the effectiveness of \textsc{RePro} on the same recycling setup as 
% ReWire~\citep{nguyen2025rewire} on DCLM. 
% Our rephraser is initialized from Qwen3-4B~\citep{yang2025qwen3}, 
% utilize GRPO algorithm~\citep{shao2024deepseekmath} to train it, 
We train our rephraser initialized from either OLMo2-1B-Instruct~\citep{walsh2025olmo} or Qwen3-4B~\citep{yang2025qwen3}, 
and
employ it to recycle a 72B organic data pool sampled from DCLM-RefinedWeb~\citep{li2024datacomp,nguyen2025rewire}. 
% following the setup of ReWire~\citep{nguyen2025rewire}.
% We then apply DCLM-fastText quality classifier to both recycled and organic data, retaining 7.2B tokens from each with the highest scores, following the identical setup as ReWire~\citep{nguyen2025rewire}.
We pretrain 400M, 1.4B, and 2.8B models on the high-quality portion of organic and recycled data.
Evaluation results on 22 downstream tasks, covering reasoning, understanding, and knowledge, demonstrate that \textsc{RePro} achieves +3.7\% to +14.5\% average accuracy (DCLM Core score) gains over the organic baseline. 
As shown in Figure~\ref{fig:overall-cmp}, \textsc{RePro} also greatly outperforms (1) the state-of-the-art web recycling method ReWire, which prompts Llama-3.3-70B-Instruct with chain-of-thought reasoning, 
and (2) organic baseline when enlarging the data pool by 4$\times$.
We also increase the amount of recycled data in final mixtures and 
% find that the optimal performance of 
% \textsc{RePro} is achieved when recycling twice the amount of high-quality organic data. 
% This means that 
find that \textsc{RePro} can boost organic data efficiency by 2-3$\times$—the amount of high-quality data in the final pretraining set relative to high-quality organic data.

% We also carefully study the effect of recycling ratio (i.e., the ratio of recycled tokens in total tokens) in different methods and 
% find that the optimal recycling ratio of \textsc{RePro} is 2-3$\times$ higher than ReWire, 

% This means that \textsc{RePro} can substantially boost the organic data efficiency by 3$\times$.
% boost the quantity of high-quality pretraining data by 3$\times$. 
% Further analyses show that our four reward functions effectively
Further ablation studies show that our reward functions effectively optimize the rephraser to generate high-quality data 
while maintaining the semantic meaning and structural diversity of organic data. 
As a representative study, we measure how many key points in one organic data are supported, omitted, or contradicted 
in its recycled version using the method from~\citet{qi2024long2rag}. 
As shown in Figure~\ref{fig:keypoint-recall}, \textsc{RePro} achieves the highest number of 95\% supported key points while significantly reducing the ratio of omitted key points by up to 92\% compared to prompting-based methods WRAP~\citep{maini2024rephrasing} and ReWire. 
In-depth analysis of rephrasing operations shows that \textsc{RePro} flexibly applies diverse operations such as paraphrasing, removing, and clarifying to enhance data quality.
These results highlight that \textsc{RePro} recycles web data in a faithful manner, 
effectively alleviating the data scarcity issue in LLM pretraining.

We summarize our main contributions as follows:
\begin{enumerate}
    \item We propose \textsc{RePro}, a novel web recycling method for better organic data efficiency by training a small LM with RL to perform effective and faithful rephrasing.
    % \item Leveraging quality and faithfulness rewards, our rephraser can effectively learn to generate high-quality data while maintaining the semantic and structure of organic data.
    % the semantic meaning and 
    % structural diversity of the original web data, preventing model collapse.
    \item We design quality and faithfulness rewards to optimize the rephraser to generate high-quality data while maintaining organic semantics, structure, and length.
    \item \textsc{RePro} doubles the gains of the state-of-the-art recycling method despite using a 70$\times$ smaller rephraser and boosts organic data efficiency by 2-3$\times$. 
    Analyses confirm that \textsc{RePro} faithfully preserves the essential semantics and structure of organic data.
\end{enumerate}
\section{Related Work}

% 1. Data synthesis for pretraining.

% 2. Learn a synthetic data generator.

% 3. Distribution collapse of synthetic data.

% 4. Data quality metrics for pretraining (not sure).

% compute-limited and data-limited regimes.

% Despite growing concern that LLM pre-training is hitting such a “data wall”, existing work on data curation still finds
% it necessary to discard the majority–sometimes up to 99\%——of the data collected to ensure quality and state-of-the-art downstream performance.
% the growth of public human-generated texts has been less sustainable.

The scaling of large language models (LLMs) has expanded along three primary dimensions: model size, compute budget, and pretraining data volume~\citep{kaplan2020scaling,Chinchilla}. 
In the early stages of LLM development, the main bottleneck was compute, which has since been largely alleviated by advances in hardware efficiency, 
novel model architectures~\citep{shazeer2017outrageously,liu2024deepseekv3}, 
and the selective use of high-quality data~\citep{engstrom2024dsdm,wettig2024qurating,yu2024mates}. 
Recent forecasts indicate that another critical constraint lies in the quantity of pretraining data, 
as the supply of organic (human-generated) text available on the internet is expected to be rapidly exhausted~\citep{villalobos2024will}. 
% This realistic challenge shifts the primary scaling bottleneck from compute to data~\citep{muennighoff2023scaling}.
% The shortage of high-quality data poses a significant challenge to the continued scaling of LLMs. 
Meanwhile, standard data curation pipelines often discard a large portion of the
collected web data to ensure the quality of pretraining~\citep{weber2024redpajama,li2024datacomp}. 
These practices highlight the scarcity of high-quality data, which has become a new bottleneck for further scaling~\citep{muennighoff2023scaling}.

% \cx{aining data in general like in midtraining and posttaining where they are very helpful, perhaps the training of data syntherzing can be discussed here. (one paragraph first), versus synthetic data in pretraining that is more challenging, that yield distribution collaposes etc (one para), then recent work in making it more conservative to simple operations and rephasing.}

\begin{figure*}
    \centering
    % \vspace{-0.2cm}
    % \vspace{-0.5cm}
    \includegraphics[width=0.8\linewidth]{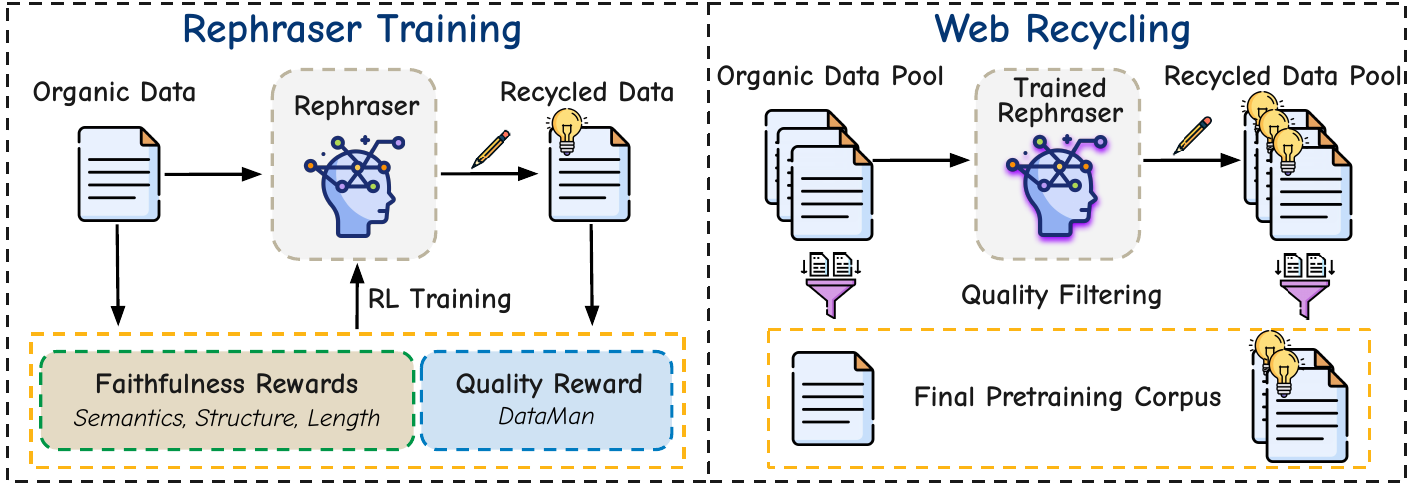}
    % \vspace{-1.5cm}
    \caption{Overview of \textsc{RePro}. We train our rephraser with quality and faithfulness rewards to optimize it to faithfully recycle the web.}
    \label{fig:pipeline}
    \vspace{-0.2cm}
\end{figure*}

As high-quality organic data becomes increasingly scarce, 
synthetic data emerges as a principled complement due to its flexibility and scalability~\citep{havrilla2024surveying}.
Recent work highlights the effectiveness of synthetic data in various training stages, spanning from pretraining~\citep{maini2024rephrasing}, mid-training~\citep{wang2025octothinker}, to post-training~\citep{li2025montessori}.
In mid- and post-training, synthetic data generation typically 
leverages a set of high-quality seeds and employs LLMs to extract its essence or produce similar examples~\citep{wang2023self}, 
which has been widely adopted to to incentivize math~\citep{ge2024scaling,zhou2025megamath}, 
reasoning~\citep{wang2025octothinker}, and instruction-following~\citep{yue2024mammoth2,li2025montessori} abilities.
In contrast, synthetic data generation for pretraining is more challenging and less understood, confronting issues such as limited diversity and reduced fidelity to real-world distributions~\citep{havrilla2024surveying}.
% due to the need to maintain diversity and coverage of the organic data~\citep{havrilla2024surveying}.
Successful attempts, including Wikipedia-style rephrasing~\citep{maini2024rephrasing}, guided rewriting~\citep{nguyen2025rewire}, 
and topic-seeded generation~\citep{li2023textbooks,allal2024cosmopedia,hao2025reformulation},  
enhance pretraining corpora in both data quantity and quality~\citep{su2025nemotroncc,abdin2024phi4,maini2025beyondweb}. 

Though promising, theory and practice caution that the 
indiscriminate use of synthetic data can trigger model collapse~\citep{shumailov2024nature,havrilla2024surveying,dohmatob2025strong}, 
degrading generalization capabilities~\citep{gerstgrasser2024inevitable,feng2025beyondcollapse}. 
Specifically, \citet{shumailov2024nature} show rapid degeneration when successive models are trained on their predecessors' outputs, 
with scaling-law changes explaining this as the tail-knowledge loss in synthetic data~\citep{dohmatob2024tale}.  
% To prevent the potential collapse, \cite{gerstgrasser2024inevitable} emphasize the mixing of real and synthetic data rather than replacement, 
% and \cite{seddik2024howbad} quantify safe mixing thresholds from a statistical perspective.
The distribution collapse issue is particularly risky, 
as it not only degrades pretraining performance but harms more on the post-training outcomes~\citep{chen2024diversity}.
To prevent the potential collapse in pretraining, ProX~\citep{zhou2024prox} and RefineX~\citep{bi2025refinex} 
restrict the refining process to conservative programs, 
cleaning pretraining data with a predefined set of operations, e.g., deletion and normalization.
\section{Methods}

In this section, we present \textsc{RePro}, 
an effective and faithful web recycling method for pretraining.
We first introduce the web recycling setup (§\ref{sec:method-recycling}) and then our reinforcement learning 
approach to optimize an LM rephraser to faithfully recycle the web (§\ref{sec:method-rl}).
Figure~\ref{fig:pipeline} illustrates our pipeline.
% Next, we describe warming up our rephraser with supervised fine-tuning (§\ref{sec:method-sft}).

\subsection{Recycling Pretraining Data}
\label{sec:method-recycling}

The construction of pretraining data starts from an organic data pool $\mathcal{D}_{\text{org}}$, obtained from web sources. As incorporating low-quality data into pretraining can significantly degrade model performance~\citep{li2024datacomp}, only high-quality subset is retained from $\mathcal{D}_{\text{org}}$. Formally, we define a quality function $Q(\cdot)$ with a threshold $\tau_{\text{org}}$ to select the high-quality subset:
\begin{align}
\mathcal{D}_{\text{org-hq}} &= \{ x \in \mathcal{D}_{\text{org}} \mid Q(x) \geq \tau_{\text{org}} \}.
\end{align}
In data-limited scenarios (e.g., training frontier LLMs~\citep{maini2025beyondweb}),
the total number of tokens in $\mathcal{D}_{\text{org-hq}}$, denoted by $B_{\text{org-hq}}$, is insufficient to cover the unique pretraining token budget $B$. To address this shortage, we introduce a language model rephraser $\mathcal{R}$, which transforms an organic data sample $x$ into a recycled data sample $x'$ given a rephrasing prompt $p$.
All rephrasings then form the recycled data pool $\mathcal{D}_{\text{rec}}$:
\begin{align}
\mathcal{D}_{\text{rec}} &= \{ x' = \mathcal{R}(p, x) \mid x \in \mathcal{D}_{\text{org}} \}.
\end{align}
As with organic data, recycled samples are filtered by $Q(\cdot)$ to select the highest-quality portion: 
\begin{align}
\mathcal{D}_{\text{rec-hq}} &= \{ x \in \mathcal{D}_{\text{rec}} \mid Q(x) \geq \tau_{\text{rec}} \},
\end{align}
where $\tau_{\text{rec}}$ is set so that the total number of tokens in $\mathcal{D}_{\text{rec-hq}}$ is $B - B_{\text{org-hq}}$.

The final pretraining dataset is then constructed from both organic and recycled high-quality subsets:
\begin{align}
\mathcal{D}_{\text{final}} &= \mathcal{D}_{\text{org-hq}} \cup \mathcal{D}_{\text{rec-hq}}.
\end{align}

\subsection{Training a Faithful Rephraser with RL}
\label{sec:method-rl}

\begin{table*}[t]
    \centering
    \caption{Reward functions used in our RL for training an effective and faithful LM rephraser.}
    \vspace{-0.6cm}
    \label{tab:reward_fn}
    \vskip 0.15in
    \resizebox{0.8\textwidth}{!}{%
    \begin{tabular}{p{1.7cm}p{1.7cm}p{4.2cm}p{6.2cm}}
        \toprule
        \textbf{Reward} & \textbf{Type} & \textbf{Functionality} & \textbf{Formal Definition} \\
        \midrule
        \textbf{\texttt{DataMan}} & Quality & Generate higher-quality data & DataMan(recycled) - DataMan(organic) \\
        \textbf{\texttt{BERTScore}} & Faithfulness & Keep semantics consistent & BERTScore(organic, recycled) $\ge \tau_{\text{BERTScore}}$ \\
        \textbf{\texttt{Structure}} & Faithfulness & Keep text structure consistent & Structure(organic, recycled) == 1 \\
        \textbf{\texttt{Length}} & Faithfulness & Penalize free-form generation & Len(recycled) $\le$ $\tau_{\text{Length}}$ $*$ Len(organic) \\
        \bottomrule
    \end{tabular}%
    }
\end{table*}

% We leverage reinforcement learning (RL) to directly optimize our rephraser for generating high-quality pretraining data, 
% while encouraging the distribution of recycled data to remain consistent with that of organic data.

% We leverage reinforcement learning (RL) to directly optimize our rephraser for generating high-quality pretraining data while faithfully preserving the characteristics of organic data.

To avoid model collapse when using synthetic data in pretraining~\citep{chen2024diversity},
we leverage reinforcement learning (RL) to train a faithful rephraser for web recycling.
Specifically, we design two types of rewards, \textit{quality} and \textit{faithfulness}, to guide RL optimization.
The quality reward encourages the rephraser to produce outputs that are of higher quality than the organic data, while faithfulness rewards ensure that the rephrased data faithfully preserves the core semantic, structure, and length of the original text. 
The specific reward functions are as follows:

\begin{itemize}
    \item \textbf{Data Quality (\texttt{DataMan}):} To encourage the generation of high-quality data, this reward incentivizes the rephraser to produce outputs $x'$ of better quality than the organic data $x$. 
    Specifically, we choose DataMan~\citep{peng2025dataman}, the state-of-the-art quality assessment method that prompts an LM to evaluate pretraining data across 13 criteria (e.g., coherence, topic focus, knowledge novelty) and assign an overall score.
    \begin{equation}
        r_{\text{DataMan}}(x, x') = \text{DataMan}(x') - \text{DataMan}(x)
    \end{equation}
    \item \textbf{Semantic Faithfulness (\texttt{BERTScore}):} To ensure that the rephrased text preserves the core meaning of the organic data, we reward semantic similarity between $x$ and $x'$ when their BERTScore~\citep{zhang2020bertscore} is higher than a preset threshold $\tau_{\text{BERTScore}}$.
    \begin{equation}
        r_{\text{\tiny{BERTScore}}}(x, x') = \mathbf{1}[\text{\scriptsize{BERTScore}}(x, x') \ge \tau_{\text{\tiny{BERTScore}}}]
    \end{equation}
    \item \textbf{Structural Faithfulness (\texttt{Structure}):} To encourage the rephraser to maintain the high-level textual structure (e.g., Markdown) of the organic data and prevent the loss of the structural diversity, we apply a structure comparison LM to assess structural faithfulness by in-context learning.
    \begin{equation}
        r_{\text{Structure}}(x, x') = \mathbf{1}[\text{Structure}(x, x') = 1]
    \end{equation}
    \item \textbf{Length Faithfulness (\texttt{Length}):} To further penalize uncontrolled or free-form generation that deviates significantly from the organic data, we impose a constraint on the length of the recycled text relative to the original by a factor of $\tau_{\text{Length}}$.
    \begin{equation}
        r_{\text{Length}}(x, x') = \mathbf{1}[\text{Len}(x') \le \tau_{\text{Length}} * \text{Len}(x)]
    \end{equation}
\end{itemize}
The final reward function $r(x, x')$ combines these individual rewards with weighting coefficients $\lambda$:
\begin{align}
    r(x, x') = &\lambda_{\text{DataMan}} r_{\text{DataMan}} + \lambda_{\text{BERTScore}} r_{\text{BERTScore}} \notag\\ &+ \lambda_{\text{Structure}} r_{\text{Structure}} + \lambda_{\text{Length}} r_{\text{Length}}.
\end{align}
We employ Group Relative Policy Optimization (GRPO) algorithm~\citep{shao2024deepseekmath} for RL training, starting from the base model $\mathcal{R}_{\text{base}}$ with organic data $x$ sampled from $\mathcal{D}_{\text{grpo}} \subset \mathcal{D}_{\text{org}}$. GRPO enhances training stability by normalizing advantage estimates from a group of $n$ generated outputs $\{x'_1, \dots, x'_n\}$. The advantage of $x'_i$ is used to update the policy via a clipped surrogate objective:

{
\vspace{-0.6cm}
\small
\begin{align}
    \mathcal{R}_{\text{rl}} &= \arg \max_{\mathcal{R}} \mathbb{E}_{\substack{x \sim \mathcal{D}_{\text{grpo}},x'_i \sim \pi_{\mathcal{R}}~}} \notag\\
    &\hspace{-2.5em}\left[ \min\left( P\hat{A}_i, \text{clip}\left(P, 1-\epsilon, 1+\epsilon\right)\hat{A}_i \right) - \beta \mathbb{D}_\text{KL}(\mathcal{R} \| \mathcal{R}_{\text{base}}) \right], \notag\\
    \text{where } P &= \frac{\pi_{\mathcal{R}}(x'_i\mid p,x)}{\pi_{\mathcal{R}_{\text{base}}}(x'_i\mid p,x)}, \notag\\\hat{A}_i &= \frac{r(x, x'_i) - \mathrm{mean}(\mathbf{r})}{\mathrm{std}(\mathbf{r})}, \notag\\\text{and } \mathbf{r} &= \{r(x, x'_1), \dots, r(x, x'_n)\}.
\end{align}
}
$\pi_{\mathcal{R}}(x'\mid p, x)$ denotes the probability of the rephraser model generating $x'$ from $x$. The final optimized model $\mathcal{R}_{\text{rl}}$ generates the recycled data pool $\mathcal{D}_{\text{rec}}$ from the organic data pool $\mathcal{D}_{\text{org}}$, which then constitutes the pretraining dataset following the steps in Section~\ref{sec:method-recycling}.

\section{Experimental Setup}
\label{sec:setup}

\begin{table*}
    [t]
    \centering
    \setlength{\tabcolsep}{3pt}
    \caption{Benchmarking different web recycling methods on DCLM with 400M, 1.4B, and 2.8B pretraining models. The ``Unique Data'' column reports the high-quality organic and rephrased token budgets as ``$B_{\text{org-hq}} + B_{\text{rec-hq}}$''; for example, ``7.2B + 7.2B'' denotes a $1{:}1$ mixture of 7.2B organic and 7.2B rephrased high-quality tokens.
    % Dependencies on stronger LLMs (e.g., LLama3-70B-Instruct) are denoted by $*$.
    % Best performances are marked \textbf{bold}.
    \textbf{Bold} and \underline{underline} indicate the best and second-best results.}
    \vspace{-0.6cm}
    \label{tab:main} 
    \vskip 0.15in
    % \begin{sc}
        \centering
        \resizebox{1.0\textwidth}{!}{%
        \begin{tabular}{l|cc|cccccc}
            \toprule                                    & & & \makecell{\textbf{Commonsense}\\\textbf{Reasoning}} & \makecell{\textbf{Language}\\\textbf{Understanding}} & \makecell{\textbf{Reading}\\\textbf{Comprehension}} & \makecell{\textbf{Symbolic}\\\textbf{Problem Solving}} & \makecell{\textbf{World}\\\textbf{Knowledge}} & \makecell{\\\textbf{Core}} \\
            \textbf{Method} &
            \makecell{\textbf{Pool}} & \makecell{\textbf{Unique Data}} & \textit{(3 tasks)}                                  & \textit{(6 tasks)}                                   & \textit{(3 tasks)}                                  & \textit{(5 tasks)}                                     & \textit{(5 tasks)}                            & \textit{(22 tasks)}        \\
            \midrule                                                        
            \multicolumn{6}{l}{\textbf{400M Setting:} 400M model, 28.8B training tokens}        \\
            \midrule
            Organic                                                                       & 72B & 7.2B & 0.23613                                             & 0.27079                                              & 0.03724                                             & 0.14535                                                & 0.20126                                       & 0.18990 \\
            Organic                                                                       & 72B & 14.4B & 0.26953 & 0.25781 & 0.05623 & 0.14991 & 0.17683 & 0.18899 \\
            WRAP                                                                       & 72B & 7.2B + 7.2B & 0.24784 & 0.25798 & 0.06269 & 0.16303 & 0.20067 & 0.19536 \\
            ProX                                                                       & 72B & 7.2B + 7.2B & 0.24252                                             & 0.25403                                              & 0.06884                                             & 0.16528                                                & 0.20647                                       & 0.19623 \\
            ReWire                                                                       & 72B & 7.2B + 7.2B & 0.24051                                             & 0.26453                                              & 0.06232                                             & 0.17392                                                & \underline{0.21246}                                       & 0.20125                    \\
            % \rowcolor{blue!10} \textsc{RePro} (w/ SFT)                                              & 72B & 7.2B + 7.2B & \textbf{0.29210}                                             & 0.25965                                     & 0.07141                                    & 0.16988                                       & \textbf{0.21632} & 0.20816 \\
            \rowcolor{blue!10} \textsc{RePro}$_{\text{\textit{OLMo2-1B}}}$                                               & 72B & 7.2B + 7.2B & \underline{0.27361} & \underline{0.27692} & \textbf{0.10292} & \underline{0.17819} & \textbf{0.22057} & \textbf{0.21749} \\
            \rowcolor{blue!10} \textsc{RePro}$_{\text{\textit{Qwen3-4B}}}$                                               & 72B & 7.2B + 7.2B & \textbf{0.28454} & \textbf{0.27792} & \underline{0.07181} & \textbf{0.19409} & 0.21154 & \underline{0.21658} \\
            \midrule
            \multicolumn{6}{l}{\textbf{1B Setting:} 1.4B model, 28.8B training tokens}        \\
            \midrule
            Organic                                                                       & 72B & 7.2B & 0.32348                                             & 0.38371                                              & 0.19584                                             & 0.19795                                                & 0.28745                                       & 0.28578                    \\
            Organic                                                                       & 72B & 14.4B & 0.31483                                             & 0.37916                                              & 0.15112                                             & 0.17507                                                & 0.28311                                       & 0.27108 \\
            WRAP                                                                       & 72B & 7.2B + 7.2B &  0.33860 & 0.36873 & 0.20071 & \underline{0.20492} & 0.27576 & 0.28335 \\
            ProX                                                                       & 72B & 7.2B + 7.2B & 0.34246                                             & 0.38253                                              &  0.21735                                             & 0.18487                                                & 0.29641                                       & 0.29004                    \\
            ReWire                                                                       & 72B & 7.2B + 7.2B & 0.33330                                             & 0.37400                                              & \underline{0.23369}                                             & 0.17641                                                & \textbf{0.31186}                                       & 0.29029                    \\

            % \rowcolor{blue!10} \textsc{RePro} (w/ SFT)                                              & 72B & 7.2B + 7.2B & 0.33996                                             & \textbf{0.38779}                                     & 0.21586                                    & 0.18217                                       & \textbf{0.31455}                                       & 0.29445 \\
            \rowcolor{blue!10} \textsc{RePro}$_{\text{\textit{OLMo2-1B}}}$                                               & 72B & 7.2B + 7.2B & \underline{0.36480} & \underline{0.38379} & \textbf{0.23370} & 0.19162 & 0.29755 & \underline{0.29746} \\
            \rowcolor{blue!10} \textsc{RePro}$_{\text{\textit{Qwen3-4B}}}$                                               & 72B & 7.2B + 7.2B & \textbf{0.36776} & \textbf{0.38519} & 0.20832 & \textbf{0.20597} & \underline{0.30304} & \textbf{0.29929} \\
            % \midrule
            % Organic                                                                       & 144B & 14.4B                                              &                                               &                                              &                                                 &                                        &                     \\
            % Organic                                                                       & 288B & 28.8B & 0.34994                                             & 0.38584                                              & 0.22059                                             & 0.18291                                                & 0.30784                                       & 0.29456                    \\
            % \midrule
            % \multicolumn{6}{l}{\textbf{3B Setting:}\footnotemark2.8B model, 28.8B training tokens}        \\
            % \midrule
            % Organic                                                                       & 72B & 7.2B & 0.39219 & 0.42597 & 0.24771 & 0.19521 & 0.32010 & 0.32055 \\
            % Organic                                                                       & 72B & 14.4B \\
            % ReWire                                                                       & 72B & 7.2B + 7.2B & \\
            % \rowcolor{blue!10} \textsc{RePro}                                                       & 72B & 7.2B + 7.2B & 0.40195 & 0.43268 & 0.24982 & 0.19823 & 0.34069 & 0.32937 \\
            % \rowcolor{blue!10} \textsc{RePro} (w/ SFT)                                               & 72B & 7.2B + 7.2B & 0.40802 & 0.42396 & 0.27067 & 0.19541 & 0.33702 & 0.32918 \\
            
            \midrule
            \multicolumn{6}{l}{\textbf{3B Setting:} 2.8B model, 55.9B training tokens}        \\
            \midrule
            Organic 
            & 144B & 14.4B 
            & 0.41490 
            & 0.47105 
            & 0.30908 
            & 0.19259 
            & 0.36491 
            & 0.35390 \\
            
            WRAP 
            & 144B & 14.4B + 14.4B 
            & 0.41334 
            & 0.44467 
            & 0.27998 
            & 0.19308 
            & 0.35327 
            & 0.33999 \\
            
            ProX 
            & 144B & 14.4B + 14.4B 
            & 0.37050 
            & 0.47198 
            & 0.31033 
            & 0.19441 
            & 0.37112 
            & 0.35688 \\
            
            ReWire 
            & 144B & 14.4B + 14.4B 
            & 0.40759 
            & 0.47302 
            & 0.31280 
            & 0.19138 
            & \underline{0.37658} 
            & 0.35632 \\
            
            \rowcolor{blue!10} \textsc{RePro}$_{\text{\textit{OLMo2-1B}}}$ 
            & 144B & 14.4B + 14.4B 
            & \textbf{0.43732} 
            & \textbf{0.48893} 
            & \underline{0.31908} 
            & \underline{0.20206} 
            & 0.37179 
            & \textbf{0.36691} \\
            
            \rowcolor{blue!10} \textsc{RePro}$_{\text{\textit{Qwen3-4B}}}$
            & 144B & 14.4B + 14.4B
            & \underline{0.41768}
            & \underline{0.47535}
            & \textbf{0.32152}
            & \textbf{0.20401}
            & \textbf{0.37801}
            & \underline{0.36272} \\

            % \rowcolor{blue!10} \textsc{RePro}
            % & 144B & 14.4B + 7.2B
            % & \textbf{0.45065}
            % & \textbf{0.48314}
            % & \textbf{0.33544}
            % & \textbf{0.20831}
            % & 0.37045
            % & \textbf{0.37050} \\

            \midrule
            \multicolumn{6}{l}{\textbf{3B Setting (Extended):} 2.8B model, 84.7B training tokens}        \\
            \midrule
            Organic
            & 144B & 14.4B
            & 0.42157
            & 0.47131
            & 0.31740
            & 0.19164
            & 0.36688
            & 0.35624 \\

            \rowcolor{blue!10} \textsc{RePro}$_{\text{\textit{OLMo2-1B}}}$
            & 144B & 14.4B + 14.4B
            & \textbf{0.44399}
            & \textbf{0.49015}
            & \textbf{0.33806}
            & \textbf{0.20736}
            & \textbf{0.37629}
            & \textbf{0.37297} \\

            \bottomrule
        \end{tabular}
        }
    % \end{sc}
    \vspace{-0.3cm}
\end{table*}
% \footnotetext{Due to compute limitations, we restrict our comparison to the best-performing organic baseline and the strongest recycling method (ReWire) in the 3B setup.}

\textbf{Pretraining Data and Model.} We conduct our main experiments on the DCLM dataset~\citep{li2024datacomp}. Specifically, we follow ReWire~\citep{nguyen2025rewire} and randomly sample 72B/144B tokens from DCLM-RefinedWeb as our organic data pool $\mathcal{D}_{\text{org}}$. 
DCLM-RefinedWeb applies rule-based filtering and global deduplication to Common Crawl~\citep{penedo2024refinedweb}, but not model-based filtering, making it a moderate-quality data source well suited for recycling.
The unique token budget $B$ is 14.4B/28.8B in our main experiment, and we explore different budgets in later analyses.
The quality function $Q$ is DCLM-fastText, the best-performing classifier in data selection from DCLM-RefinedWeb.
Following the quality filtering threshold $\tau_{\text{org}}$=0.018112 in DCLM, the amount of high-quality organic data $B_{\text{org-hq}}$ is 7.2B/14.4B.
Symmetrically, the rephrased high-quality budget is set to $B_{\text{rec-hq}} = B_{\text{org-hq}}$, so the final pretraining mixture in our main experiments is exactly 1:1 organic-to-rephrased.
For our pretraining models, we use a decoder-only architecture and adopt three commonly used parameter scales (400M, 1.4B, and 2.8B) from previous works~\citep{yu2025group,nguyen2025rewire}, and train all models from scratch.

\textbf{Evaluation.} We evaluate pretrained models on 22 downstream tasks in either zero-shot or few-shot manners. These tasks provide
a comprehensive assessment of essential model abilities, including
commonsense reasoning, language understanding, reading comprehension, symbolic problem
solving, and world knowledge. We use centered accuracy as the primary metric, where accuracy per task is mapped to 0 for random guessing and 1 for perfect accuracy. The average centered accuracy across all tasks is denoted as ``Core score''. Following~\citet{yu2025group}, we exclude CommonsenseQA from the original DCLM-Core due to its high instability and limited informativeness. Full task details and their few-shot numbers are provided in Appendix~\ref{sec:app-tasks}.

\textbf{Baselines.} We compare \textsc{RePro} with (1) \textit{organic data only};
(2) \textit{WRAP}~\citep{maini2024rephrasing}: Wikipedia-style rephrasing using Mistral-7B-Instruct-v0.1~\citep{jiang2023mistral7b}; (3) \textit{ProX}~\citep{zhou2024prox}: fine-grained edits of organic data via model-generated programs; and (4) \textit{ReWire}~\citep{nguyen2025rewire}: guided rewriting with chain-of-thought reasoning using Llama-3.3-70B-Instruct~\citep{dubey2024llama}. As ReWire's code has not been open-sourced, we randomly sample 7.2B tokens from its released data. This may give ReWire a performance advantage, since it essentially utilizes a larger pool (our organic + their organic).
These baselines cover state-of-the-art web recycling techniques such as prompting, program-based editing, and chain-of-thought reasoning.
Some concurrent works, such as RefineX~\citep{bi2025refinex}, % and BeyondWeb~\citep{maini2025beyondweb}, 
have not open-sourced their data and code, so we leave the comparison with them in the future. More baseline details can be found in Appendix~\ref{sec:app-baselines}.

\textbf{Implementation Details.} 
We initialize our rephraser model $\mathcal{R}$ with either \textit{OLMo2-1B-Instruct}~\citep{walsh2025olmo} or \textit{Qwen3-4B}~\citep{yang2025qwen3}.
OLMo2-1B-Instruct is a fully open, academically released model, while Qwen3-4B is a stronger industry-released model; together, they allow us to evaluate the generalization of our method across different model families and sizes.
In RL, we set the cutting threshold $\tau_{\text{BERTScore}}$ as 0.65, and set $\tau_{\text{Length}}$ as 1.25.
We utilize released DataMan (1.5B) and BERTScore (350M) models for their reward calculation and construct our structure comparison model by prompting Qwen3-4B with few-shot examples.
The coefficients $\lambda_{\text{DataMan}}$, $\lambda_{\text{BERTScore}}$, $\lambda_{\text{Structure}}$, and $\lambda_{\text{Length}}$ are set to 3, 1, 1, 1, respectively.
The size of the RL dataset $\mathcal{D}_{\text{grpo}}$ is 41k.
$\mathcal{D}_{\text{grpo}}$ consists solely of organic data with a DataMan score below 5, since 5 is the maximum score and thus cannot be further improved.
The clipping $\epsilon$ in GRPO is 0.2, $\beta$ is 0.005, and the number of rollouts $n$ per input is 8.
To examine whether a strong prior can improve RL, we also add an \textbf{optional} supervised fine-tuning (SFT) stage before RL, where we use GPT-4o~\citep{achiam2023gpt} as the teacher to generate 50k example rephrasings to warm up our rephraser. 
Unless otherwise stated, \textsc{RePro} denotes \textbf{RL w/o SFT}.
In the inference stage, we adopt \texttt{vllm}~\citep{kwon2023efficient} framework for efficient text generation using our rephraser. Following~\citet{nguyen2025rewire}, we use a temperature of 1.0 and a top-p of 0.9.
We conduct a key hyperparameter analysis in Appendix~\ref{sec:app-hyper} and provide all prompts used in our experiments in Appendix~\ref{sec:app-prompts}.
\section{Evaluation Results}
\begin{table*}
  [t]
  \setlength{\tabcolsep}{3pt}
  \caption{Ablation studies for
    \textsc{RePro}$_{\text{\textit{Qwen3-4B}}}$ in the 400M setup, alongside a
    non-rephrasing baseline (Nemotron-CC Diverse QA). Prompting$^+$ and SFT$^+$ denote the
    use of an enhanced prompt (Appendix~\ref{sec:app-prompts}) with reward descriptions for
  a fairer comparison.}
  \vspace{-0.6cm}
  \label{tab:ablation}
  \vskip 0.15in
  % \begin{sc}
  \centering
  \resizebox{1.0\textwidth}{!}{%
    \begin{tabular}{l|cc|cccccc}
      \toprule                                    & & &
      \makecell{\textbf{Commonsense}\\\textbf{Reasoning}} &
      \makecell{\textbf{Language}\\\textbf{Understanding}} &
      \makecell{\textbf{Reading}\\\textbf{Comprehension}} &
      \makecell{\textbf{Symbolic}\\\textbf{Problem Solving}} &
      \makecell{\textbf{World}\\\textbf{Knowledge}} & \makecell{\\\textbf{Core}} \\
      \textbf{Method} &
      \makecell{\textbf{Pool}} & \makecell{\textbf{Unique Data}} & \textit{(3 tasks)}
      & \textit{(6 tasks)}                                   & \textit{(3 tasks)}
      & \textit{(5 tasks)}                                     & \textit{(5 tasks)}
      & \textit{(22 tasks)}        \\
      \midrule
      \multicolumn{6}{l}{\textbf{400M Setting:} 400M model, 28.8B training tokens}        \\
      \midrule
      Organic                                                                       & 72B
      & 14.4B & 0.26953 & 0.25781 & 0.05623 & 0.14991 & 0.17683 & 0.18899 \\
      Prompting                                               & 72B & 7.2B + 7.2B &
      0.24310 & 0.26758 & 0.05075 & 0.17392 & 0.20196 & 0.19847          \\
      Prompting$^+$ & 72B & 7.2B + 7.2B & 0.27831 & 0.25229 & 0.06870 & 0.16475 & 0.20032
      & 0.19910 \\
      SFT                                               & 72B & 7.2B + 7.2B & 0.24447 &
      0.24920 & 0.04013 & 0.16564 & 0.21009 & 0.19216           \\
      SFT$^+$ & 72B & 7.2B + 7.2B & 0.26049 & \underline{0.27111} & 0.06593 & 0.16899 &
      0.20206 & 0.20278 \\
      RL w/o Faithfulness                                               & 72B & 7.2B +
      7.2B & 0.22357 & 0.25633 & 0.06822 & 0.16961 & 0.20376 & 0.19456 \\
      SFT + Full RL                                              & 72B & 7.2B + 7.2B &
      \textbf{0.29210}                                             & 0.25965
      & \underline{0.07141}                                    & 0.16988
      & \textbf{0.21632} & 0.20816 \\
      Full RL (No Recycling from $\mathcal{D}_{\text{org-hq}}$)
      & 72B & 7.2B + 7.2B & \underline{0.28674} & 0.27051 & 0.06737 & \underline{0.17920}
      & 0.20992 & \underline{0.21050}           \\
      \rowcolor{blue!10} Full RL (\textsc{RePro})
      & 72B & 7.2B + 7.2B & 0.28454 & \textbf{0.27792} & \textbf{0.07181} &
      \textbf{0.19409} & \underline{0.21154} & \textbf{0.21658}           \\
      \midrule
      Diverse QA (Qwen3-30B-A3B)                                                    & 72B
      & 7.2B + 7.2B & 0.24076 & 0.26523 & 0.11184 & 0.15748 & 0.22993 & 0.20847 \\
      \bottomrule
    \end{tabular}
  }
  % \end{sc}
\end{table*}

In this section, we present our main results on DCLM,
along with different unique token budgets (§\ref{sec:results-main}).
Then, we perform ablation studies (§\ref{sec:results-ablation}),  analyze the recycling
patterns of non-high-quality data (§\ref{sec:results-recycling}),
% dive into the RL training dynamics with different reward choices (§\ref{sec:results-rl-training}),
and validate the effectiveness of our reward functions (§\ref{sec:results-distribution}).
% We conclude with an in-depth analysis of the operations performed by our rephraser
% (§\ref{sec:results-operations}).
Additional results and ablations can be found in Appendix~\ref{sec:app-results}.

% Finally, we explore different quality reward options to validate our design choice
% (§\ref{sec:results-quality}).

\subsection{Main Results}
\label{sec:results-main}

\paragraph{Overall Performance.} Table~\ref{tab:main} demonstrates the overall
performance of different web recycling methods on DCLM-RefinedWeb.
% First, using top-10\% high-quality organic data outperforms using top-20\%, confirming
% the necessity of quality filtering.
Across three setups, \textsc{RePro} significantly outperforms using only organic data,
achieving 3.7\%-14.5\% relative improvements in Core scores\footnote{We discuss detailed
per-category performance in Appendix~\ref{sec:app-per-category}.}; these gains exceed two
standard deviations of the organic baseline across three random seeds in every category
(see Appendix~\ref{sec:app-variance}).
This indicates that our recycled data can
effectively complement organic data in the data-limited setup to benefit pretraining.
Our method also consistently \textbf{doubles the performance gains} achieved by all
baselines across different setups,
including the one (ReWire) that prompts Llama-3.3-70B-Instruct to rephrase the corpus.
The efficiency advantage of \textsc{RePro} is shown in Appendix~\ref{sec:app-efficiency},
where it achieves up to 87.2$\times$ speedup compared to ReWire. Surprisingly, using
OLMo2-1B as the initialization model for \textsc{RePro} achieves comparable, and in some
cases better, gains than using Qwen3-4B, demonstrating the robustness of our method
across different base rephrasers. We hypothesize that, although Qwen3 exhibits stronger
general capabilities, OLMo2 may offer greater flexibility for adaptation to the
rephrasing task; recent work likewise identifies $\sim$1B as a sweet spot for rephraser
size~\citep{niklaus2026_the_synthetic_data_playbook}.

We would like to note that the 3B setup follows the DCLM 3B-1x convention (2.8B model,
55.9B tokens) and is less saturated than 400M runs. Therefore, we continue to pretrain
both the organic baseline and \textsc{RePro} for an additional 28.8B tokens (84.7B total,
last block of Table~\ref{tab:main}) actually grows the gap from +3.7\% to +4.7\%,
confirming that the trend extends to longer training. We also observe that in the 3B
setup, WRAP can even harm performance, likely due to its limited diversity and weaker
preservation of the original distribution, both of which become increasingly important
for larger pretraining models.
% With SFT prior, \textsc{RePro} may achieve further gains on world knowledge tasks but
% falls behind in the overall Core score.
% We hypothesize that the SFT data generated by GPT-4o may contain distillation-style rephrasings
% that benefit knowledge-intensive tasks but hinder generalization to others,
% a common issue noted in previous works~\citep{su2025nemotroncc,maini2025beyondweb}.

In summary, these results highlight that even a small model
can learn without the guidance of an LLM to be an effective rephraser to generate
high-quality recycled data.

% \begin{wrapfigure}
% {r}{0.428\textwidth}
\begin{figure}
  \centering
  % \vspace{-0.1cm}
  \includegraphics[width=0.75\linewidth]{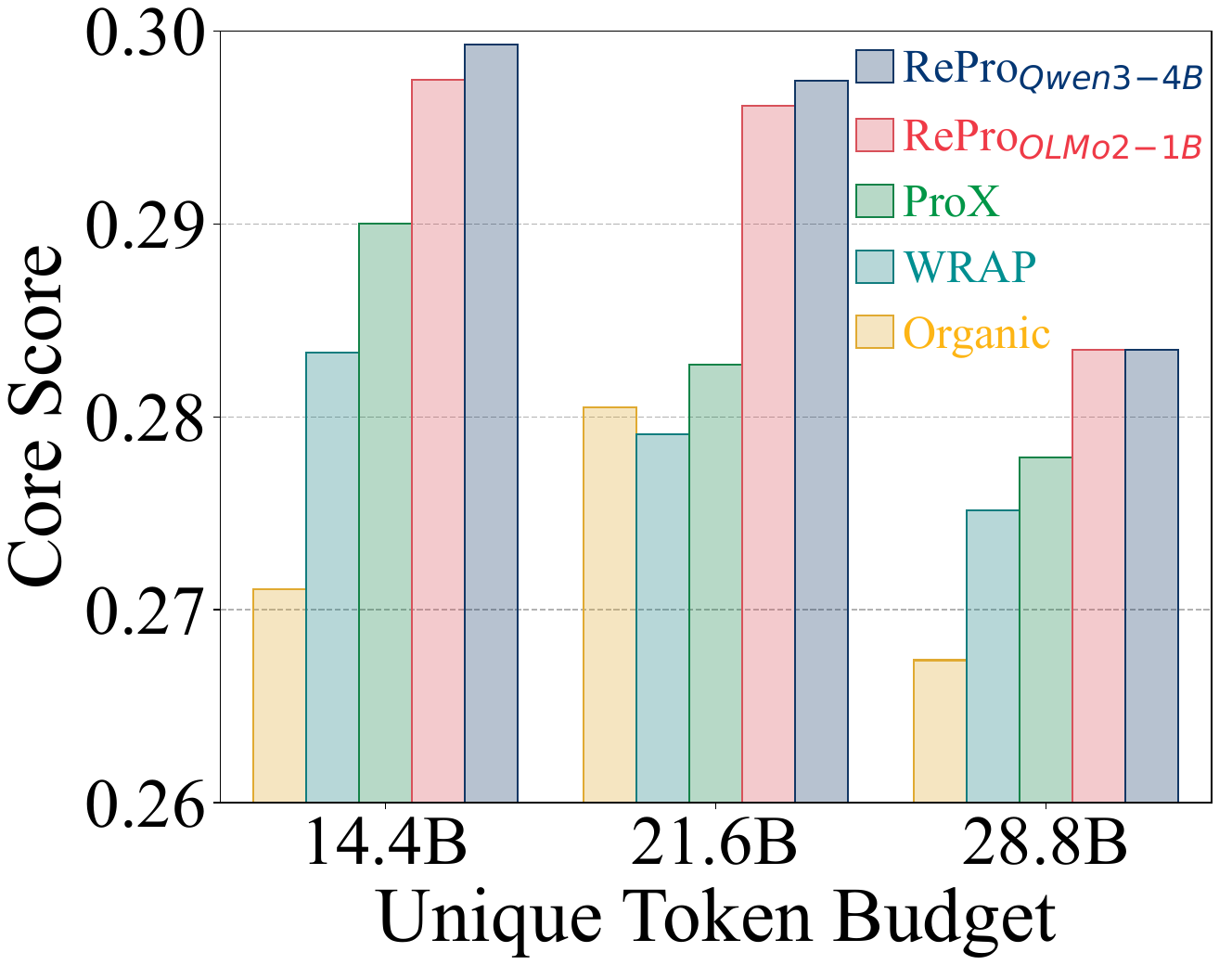}
  % \vspace{-0.6cm}
  \caption{Performance of recycling methods w.r.t. different unique token budgets in the
    1B setup. We exclude ReWire from this comparison because it does not release the full
  recycled pool $\mathcal{D}_{\text{rec}}$, and reconstructing it is beyond our compute capacity.}
  \label{fig:unique}
  \vspace{-0.6cm}
\end{figure}
% \end{wrapfigure}

\begin{figure*}[t]
  \centering
  % \vspace{-0.3cm}
  \begin{subfigure}[t]
    {0.237\textwidth}
    \includegraphics[width=1.0\linewidth]{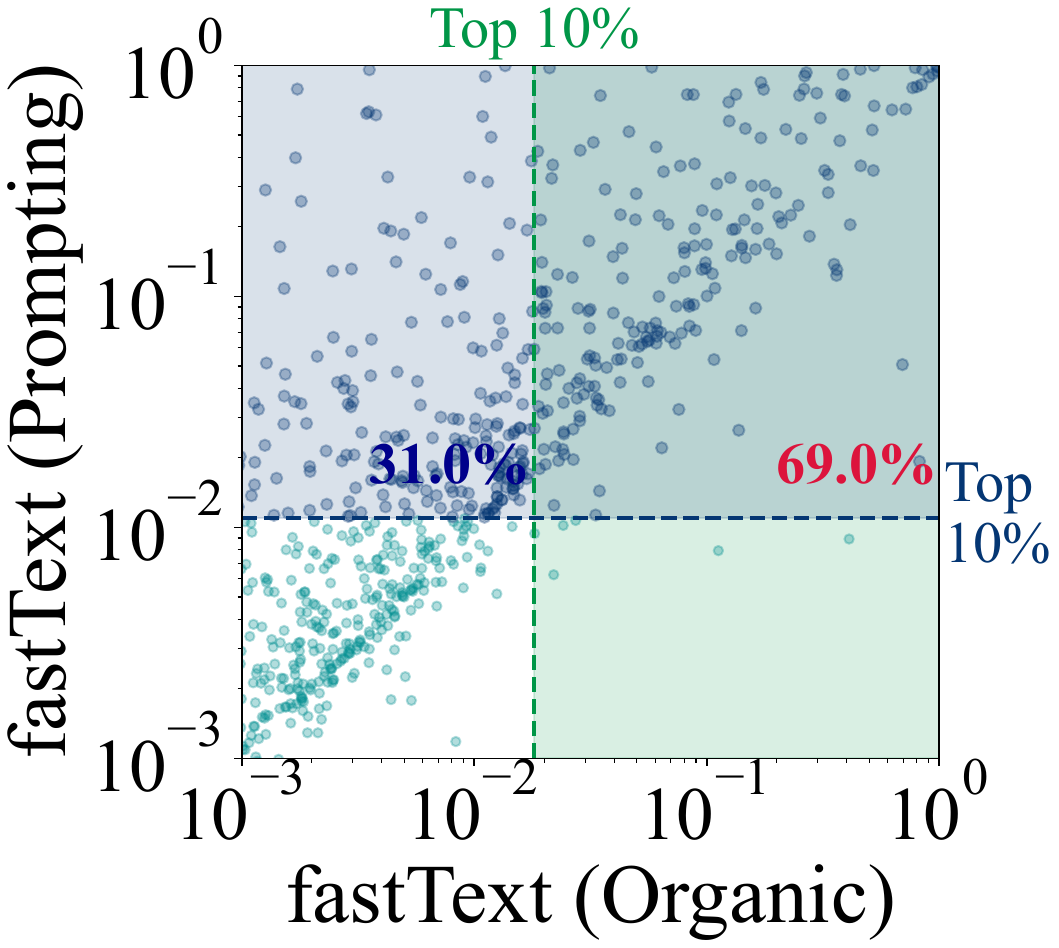}
    \caption{Prompting (Qwen3-4B)}
    \label{fig:fasttext-prompting}
  \end{subfigure}
  ~
  \begin{subfigure}[t]
    {0.237\textwidth}
    \centering
    \includegraphics[width=1.0\linewidth]{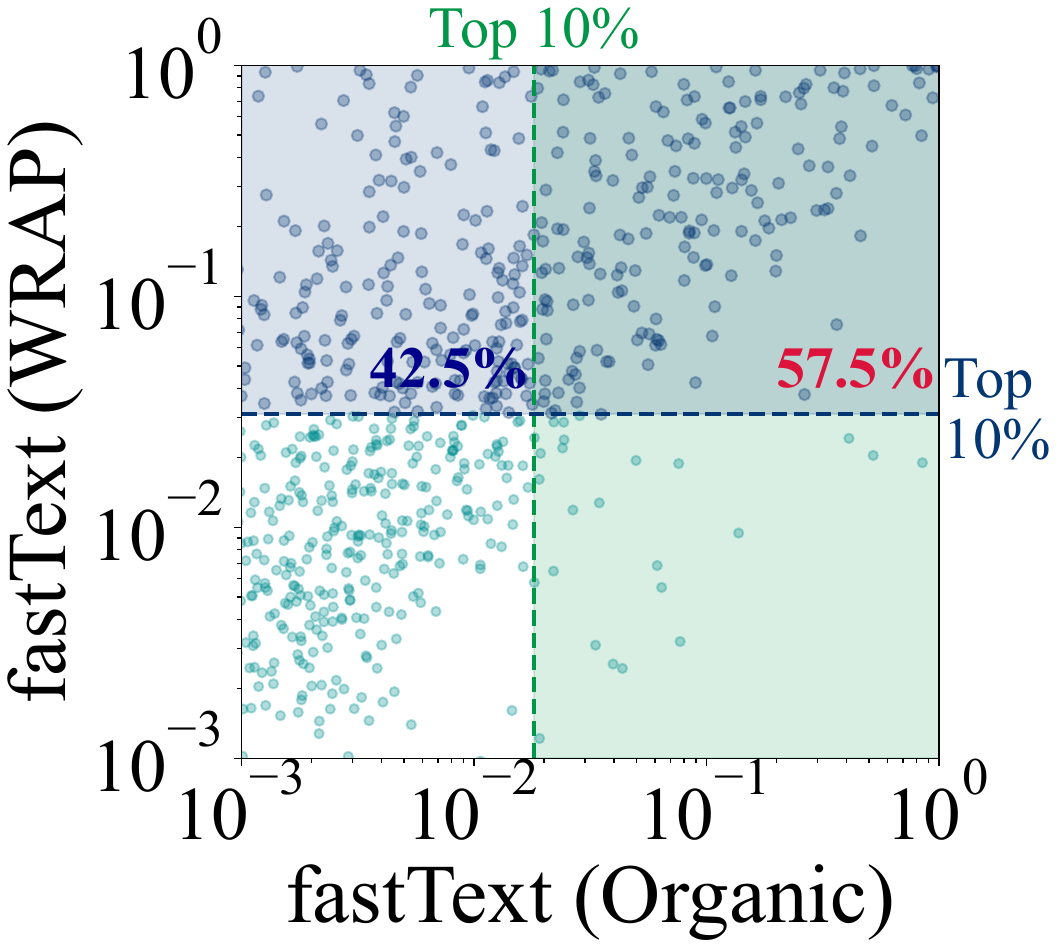}
    \caption{WRAP}
    \label{fig:fasttext-wrap}
  \end{subfigure}
  ~
  \begin{subfigure}[t]
    {0.237\textwidth}
    \centering
    \includegraphics[width=1.0\linewidth]{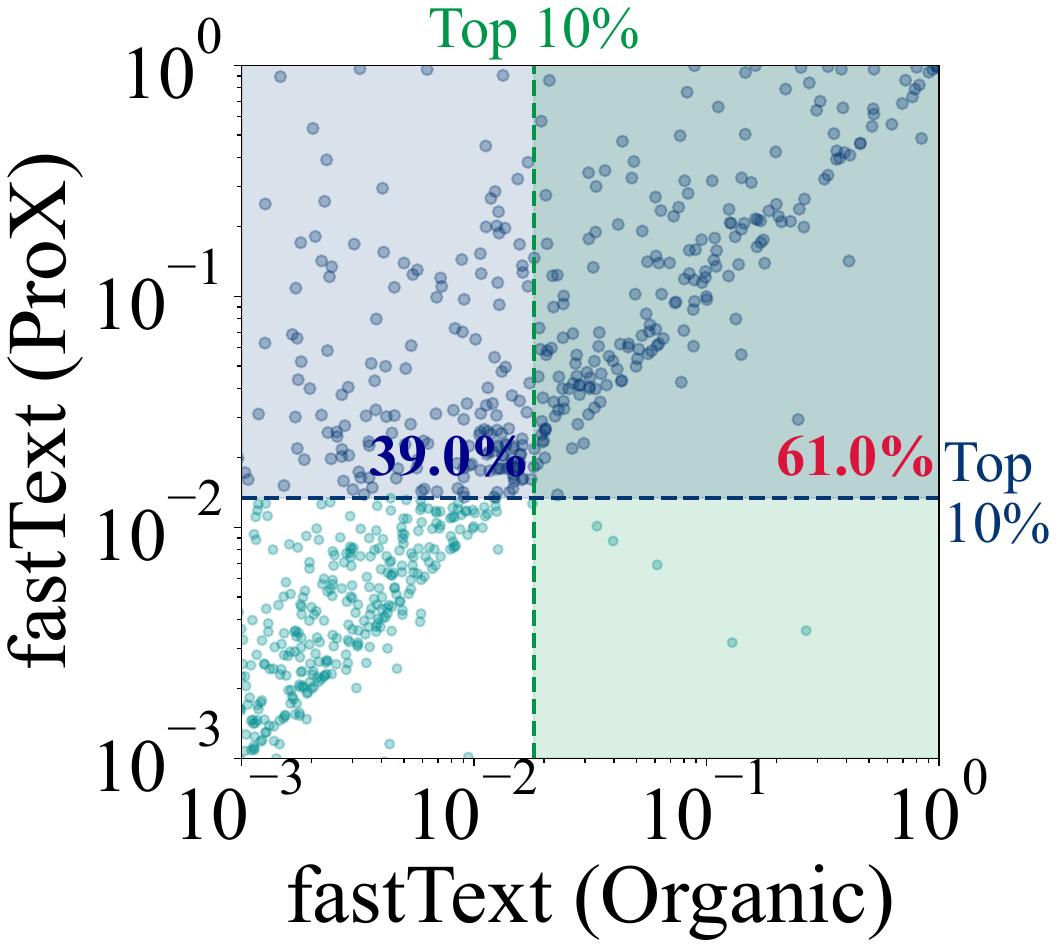}
    \caption{ProX}
    \label{fig:fasttext-prox}
  \end{subfigure}
  ~
  \begin{subfigure}[t]
    {0.237\textwidth}
    \centering
    \includegraphics[width=1.0\linewidth]{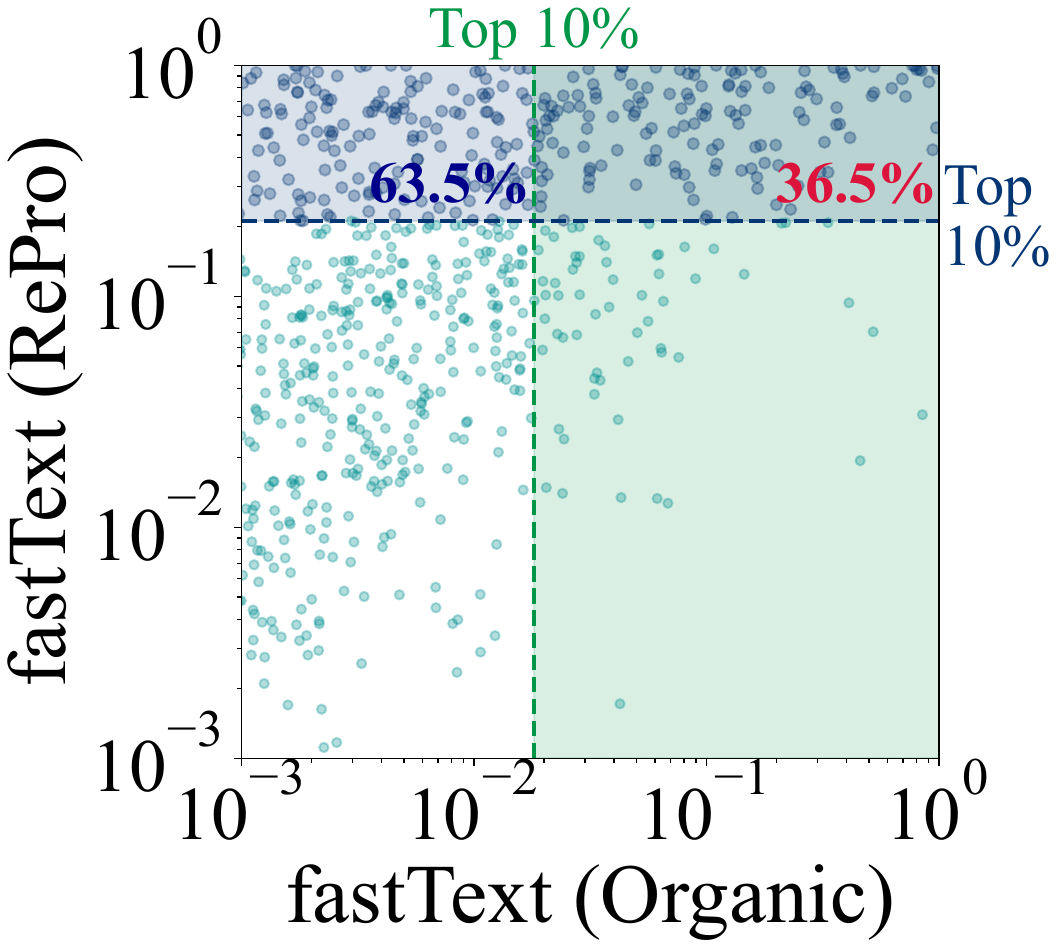}
    \caption{\textsc{RePro}$_{\text{\textit{Qwen3-4B}}}$}
    \label{fig:fasttext-repro}
  \end{subfigure}
  \caption{Distribution of fastText scores before and after each rephrasing method. The
    left bar shows $P_l$ (Eq.~\ref{eq:pl}), the fraction of high-quality recycled examples
    whose source was low-quality organic data, and the right bar shows its complement $P_r$
  (Eq.~\ref{eq:pr}).}
  \label{fig:fasttext-scatters}
  \vspace{-0.2cm}
\end{figure*}

\paragraph{Improving Organic Data Efficiency.}
We also evaluate the effectiveness of \textsc{RePro} under different unique token budgets $B$.
% seeking its upper-bound performance given a fixed organic data pool.
Specifically, we vary $B$ from 14.4B in our main experiment to 21.6B and 28.8B (our total
training budget) in the 1B setup.
% For \textsc{RePro}, we fix the amount of high-quality organic tokens
% $B_{\text{org-hq}}$ as 7.2B and vary the amount of recycled data to match the total budget.
As shown in Figure~\ref{fig:unique}, \textsc{RePro} consistently outperforms all
baselines across different $B$ by a large margin.
Among these budgets, the best performance of \textsc{RePro} is achieved at $B$=14.4B, and
$B$=21.6B yields comparable performance but delivers the largest improvement over baseline methods.
This implies that \textsc{RePro} can potentially improve organic data efficiency by 2-3$\times$.
Further increasing $B$ to 28.8B leads to a performance drop, likely due to the inclusion
of more moderate-to-low quality data, which reduces the advantage of higher quality recycled tokens.
Overall, these results highlight the effectiveness of our method in increasing the amount
of high-quality pretraining data.
% recycling large amounts of web data into high-quality training data,
% thereby significantly improving LLM pretraining performance.
% The full evaluation results of this study are provided in Table~\ref{tab:unique}.

% \paragraph{Varying Recycling Ratio.} In this experiment,
% we vary the recycling ratio $\alpha$ from 0.25 to 1.0 in the 1B setup with a fixed
% total budget of 28.8B tokens. TBD.

\subsection{Ablation Studies}
\label{sec:results-ablation}

\begin{figure*}
  \centering
  % \vspace{-0.4cm}
  \includegraphics[width=0.75\linewidth]{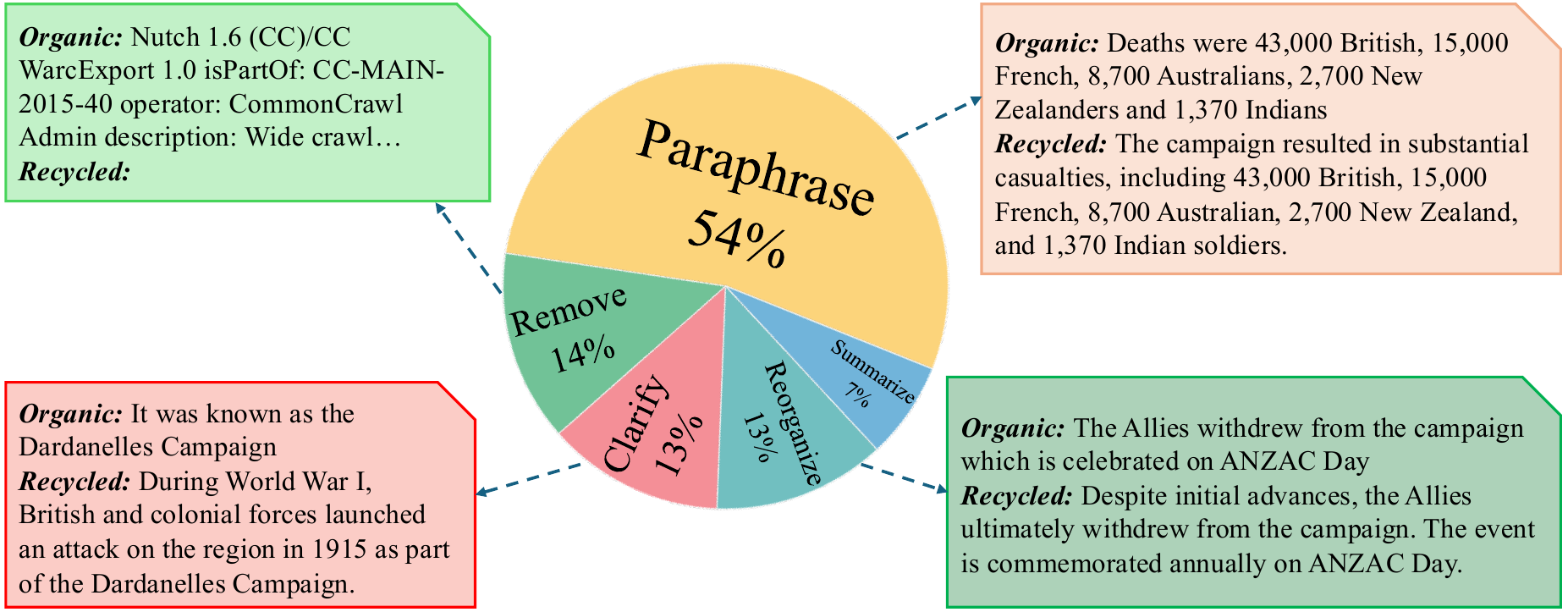}
  % \vspace{-1.5cm}
  \caption{Operations performed by \textsc{RePro} and their corresponding cases.}
  \label{fig:operations}
  \vspace{-0.2cm}
\end{figure*}

\begin{figure*}[t]
  \centering
  % \vspace{-0.3cm}
  \begin{subfigure}[t]
    {0.224\textwidth}
    \centering
    \includegraphics[width=1.0\linewidth]{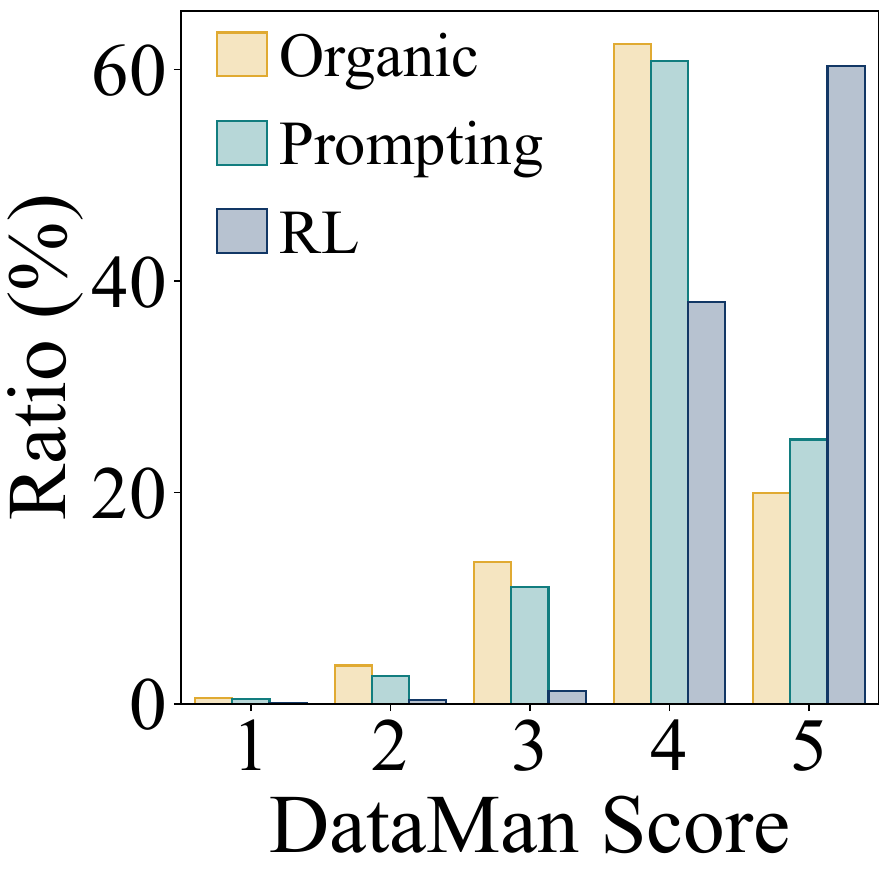}
    \caption{DataMan}
    \label{fig:ablation-distribution-dataman}
  \end{subfigure}
  ~
  \begin{subfigure}[t]
    {0.226\textwidth}
    \centering
    \includegraphics[width=1.0\linewidth]{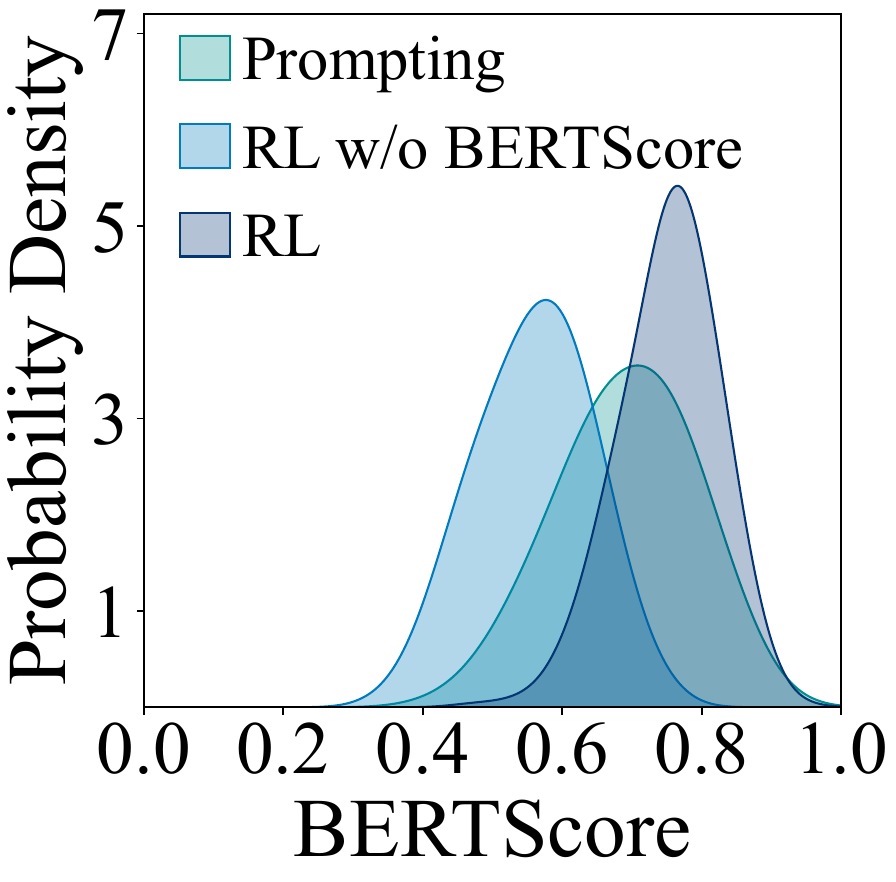}
    \caption{BERTScore}
    \label{fig:ablation-distribution-bertscore}
  \end{subfigure}
  ~
  \begin{subfigure}[t]
    {0.233\textwidth}
    \raisebox{-8pt}
    {\includegraphics[width=1.0\linewidth]{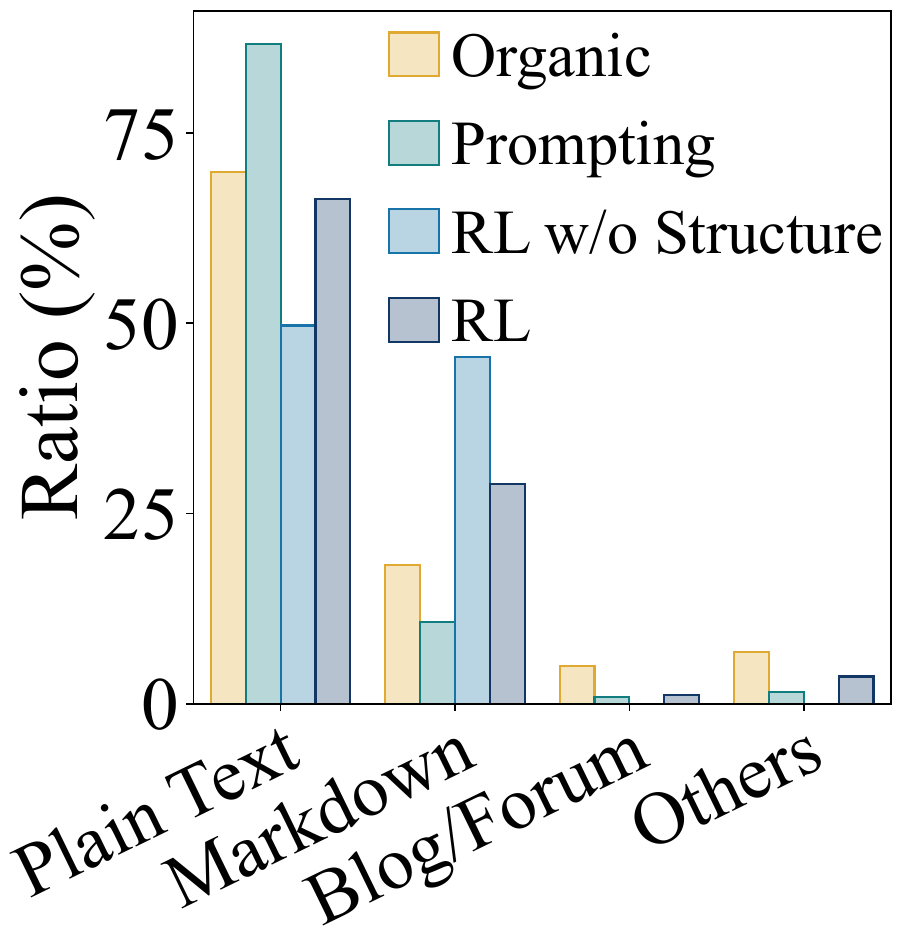}}
    \vspace{-1.1em}
    \caption{Structure}
    \label{fig:ablation-distribution-structure}
  \end{subfigure}
  ~
  \begin{subfigure}[t]
    {0.250\textwidth}
    \centering
    \includegraphics[width=1.0\linewidth]{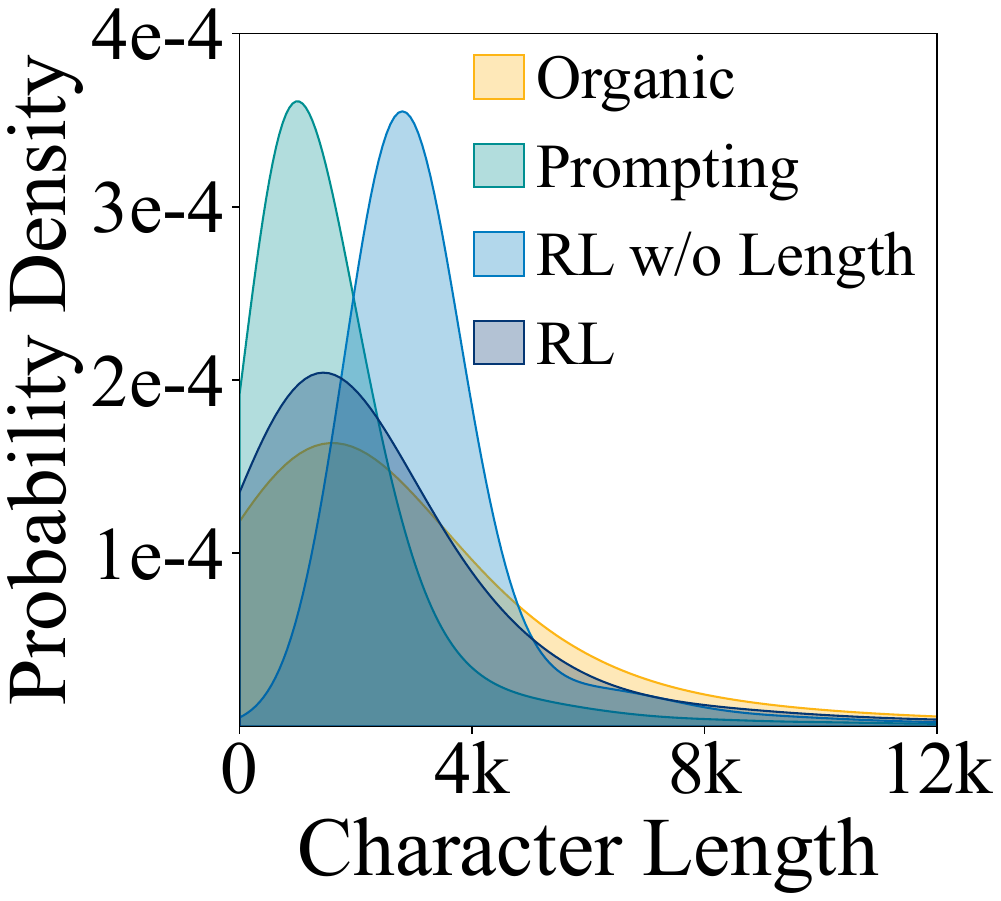}
    \caption{Length}
    \label{fig:ablation-distribution-len}
  \end{subfigure}
  \caption{Reward-related feature distributions of 30,000 recycled data generated by
    different rephrasers (Qwen3-4B as base rephraser). Features are (a) DataMan score, (b)
  BERTScore, (c) structure type, and (d) text length.}
  \label{fig:ablation-distribution}
  \vspace{-0.2cm}
\end{figure*}

We conduct ablation studies in the 400M setup to investigate the effectiveness of each
component in our rephraser training.
Comparisons include our base rephraser (Qwen3-4B) with direct prompting,
the rephraser after SFT using GPT-4o rephrasings,
and the rephraser from RL training without faithfulness rewards.
% Note that we do not include the model trained without the quality reward,
% as in this case, the rephraser would simply learn to copy the organic data to maximize
% faithfulness rewards.
We exclude the rephraser trained without the quality reward, as it would simply learn to
copy the organic data to maximize faithfulness.
As compared in Table~\ref{tab:ablation},
straightforward prompting or SFT alone could not greatly benefit overall performance
compared to the organic baseline.
% despite a notable gain on world knowledge tasks.
This implies a significant performance gap between prompting-based rephrasers
(even as strong as GPT-4o) and our training-based rephraser,
which has been specifically optimized for generating better recycled data.

With SFT prior, RL may achieve further gains on world knowledge tasks but falls behind in
the overall Core score.
We hypothesize that the SFT data generated by GPT-4o may contain distillation-style rephrasings
that benefit knowledge-intensive tasks but hinder generalization to others,
a common issue noted in previous works~\citep{su2025nemotroncc,maini2025beyondweb}.
% This comparison highlights that \textsc{RePro} does not rely on supervised signals from
% an external LLM to perform effective recycling.
Removing faithfulness rewards leads to a substantial performance drop, indicating that
the faithfulness of the rephraser is also crucial to the effectiveness of recycling.

To make the comparison with our method more informative, we also strengthen prompting and
SFT baselines by adding the descriptions of our reward functions directly into the
prompt, denoted as Prompting$^+$ and SFT$^+$. The results show that Prompting$^+$ yields
only a negligible improvement over our base prompt, and SFT$^+$, although better than
SFT, still remains clearly below \textsc{RePro}. These results demonstrate that a
dedicated RL procedure is more effective than prompting alone for shaping the base model
into a faithful and high-performing rephraser.

We further run experiments to verify whether the empirical gains of \textsc{RePro} arise
from recycling low-quality organic data. Specifically, we construct
$\mathcal{D}_{\text{rec-hq}}$ only from $\mathcal{D}_{\text{org}} \setminus
\mathcal{D}_{\text{org-hq}}$ and perform an ablation titled ``No Recycling from
$\mathcal{D}_{\text{org-hq}}$.'' Even when excluding $\mathcal{D}_{\text{org-hq}}$ from
the recycling process, we recover \textbf{77.2\%} of the total gains. However, in
realistic settings, it is more practical to utilize the full
$\mathcal{D}_{\text{rec-hq}}$ for better performance, since high-quality data can also be
transformed to improve learnability.

To further isolate the benefit of \emph{faithful rephrasing} over \emph{generative}
synthetic data, we add a Nemotron-CC Diverse QA~\citep{su2025nemotroncc} baseline that
prompts Qwen3-30B-A3B on $\mathcal{D}_{\text{org-hq}}$ to generate question-answer pairs,
and merges the generated data with $\mathcal{D}_{\text{org-hq}}$. We find that Diverse QA
helps on reading comprehension and world knowledge (QA-style tasks) but lags
\textsc{RePro} on the overall Core score despite using a much stronger generator. As
Diverse QA imposes \emph{no faithfulness constraint}, much of its gain likely comes from
distilling the generator's parametric knowledge rather than from improving the organic
distribution, which is qualitatively different from \textsc{RePro}'s recycling.

% In summary, our ablation studies highlight the necessity of using RL and ensuring
% faithfulness in our rephraser training.

\subsection{Recycling of Non-High-Quality Data}
\label{sec:results-recycling}

In this experiment, we measure how much of the high-quality recycled data $\mathcal{D}_{\text{rec-hq}}$ comes from non-high-quality organic data $\mathcal{D}_{\text{org}}\setminus\mathcal{D}_{\text{org-hq}}$.
We exclude ReWire from this comparison because its high-quality cutoff threshold is not publicly available.
Formally, we partition $\mathcal{D}_{\text{rec-hq}}$ by whether the source organic example $x$ is low- or high-quality:
\begin{align}
P_l &= \frac{|\{x' \in \mathcal{D}_{\text{rec-hq}} \mid x \in \mathcal{D}_{\text{org}}\setminus\mathcal{D}_{\text{org-hq}}\}|}{|\mathcal{D}_{\text{rec-hq}}|}, \label{eq:pl} \\
P_r &= 1 - P_l = \frac{|\{x' \in \mathcal{D}_{\text{rec-hq}} \mid x \in \mathcal{D}_{\text{org-hq}}\}|}{|\mathcal{D}_{\text{rec-hq}}|}. \label{eq:pr}
\end{align}
$P_l$ (left bar in Figure~\ref{fig:fasttext-scatters}) is the proportion of high-quality recycled examples whose source is low-quality organic data, and $P_r$ is its complement.

As illustrated in Figure~\ref{fig:fasttext-scatters}, for \textsc{RePro}, $P_l$ is almost \textbf{two-thirds (63.5\%)} of $\mathcal{D}_{\text{rec-hq}}$. In comparison, the corresponding proportions for Prompting, WRAP, and ProX are 31.0\%, 42.5\%, and 39.0\%, respectively, indicating that \textsc{RePro} converts substantially more low-quality data into high-quality rephrasings. Furthermore, \textsc{RePro}'s top-10\% fastText cutoff exceeds \textbf{0.2}, while the best baseline only reaches around \textbf{0.03}, suggesting that our rephraser achieves a substantially larger improvement in the overall fastText scores. This further explains why \textsc{RePro} demonstrates better organic data efficiency in Figure~\ref{fig:unique}. In Appendix~\ref{sec:app-hq-only}, we show that training only on this high-quality recycled portion $\mathcal{D}_{\text{rec-hq}}$ already outperforms training only on high-quality organic data, though the best results still come from merging the two, indicating that the recycled data serves as a strong \emph{complement} to organic data rather than a replacement.

We note that the fastText classifier used in this analysis is \emph{out-of-distribution} relative to the DataMan reward optimized during RL training. Quality gains that show up on fastText therefore cannot be artifacts of overfitting to the DataMan reward signal. We choose DataMan as the RL quality reward because it is a rubric-based LLM-as-a-Judge that scores text on 13 criteria (e.g., coherence, topic focus, knowledge novelty); such multi-criteria rubric rewards have recently been shown to be more robust signals for RL than single-axis scores~\citep{gunjal2026rubrics}.

% \subsection{Rephrasing Operations}
% \label{sec:results-operations}

To better understand how \textsc{RePro} operates on different data quality, 
% we provide more recycling cases in Appendix~\ref{sec:case-repro}. Thanks to the RL optimization, \textsc{RePro} preserves the key characteristics of organic data (writing style, structure, genre, etc.) and learns to flexibly apply various operations to polish the text. For instance, \textsc{RePro} often filters out noisy or irrelevant content, clarifies unclear information, and refines the organization and writing of already high-quality texts to further enhance their learnability.
% In summary, our method effectively transforms diverse data into higher-quality forms, enabling more efficient use of organic data.
we analyze the specific operations performed by our rephraser. 
Here, we randomly sample 100 instances from our recycled data pool and 
again prompt \textit{Gemini 2.5 Flash-Lite} to 
classify the operations performed by our rephraser.
We categorize the extracted operations into 5 primary types: 
\textit{paraphrasing}, \textit{removing}, \textit{clarification}, \textit{reorganization}, and \textit{summarization}.
The distributions and examples of 
each operation are illustrated in Figure~\ref{fig:operations}.
We observe that paraphrasing is the most popular operation, which can improve writing quality 
by fixing grammatical errors and enhancing fluency.
Removing is the second common one, 
which eliminates irrelevant content such as advertisements and metadata.
Other operations like clarification, reorganization, and summarization
are also frequently used to enhance the informativeness and coherence of the text.
Overall, these results demonstrate that our rephraser learns to flexibly apply a variety of operations 
(instead of a manually defined set) to
effectively recycle web data into high-quality training data. We provide more recycling cases in Appendix~\ref{sec:case-repro}.

\subsection{Effectiveness of Adopted Reward Functions}
\label{sec:results-distribution}

\begin{figure*}[t]
  \centering
  % \vspace{-0.8cm}
  \begin{subfigure}[t]
    {0.24\textwidth}
    \centering
    \includegraphics[width=1.0\linewidth]{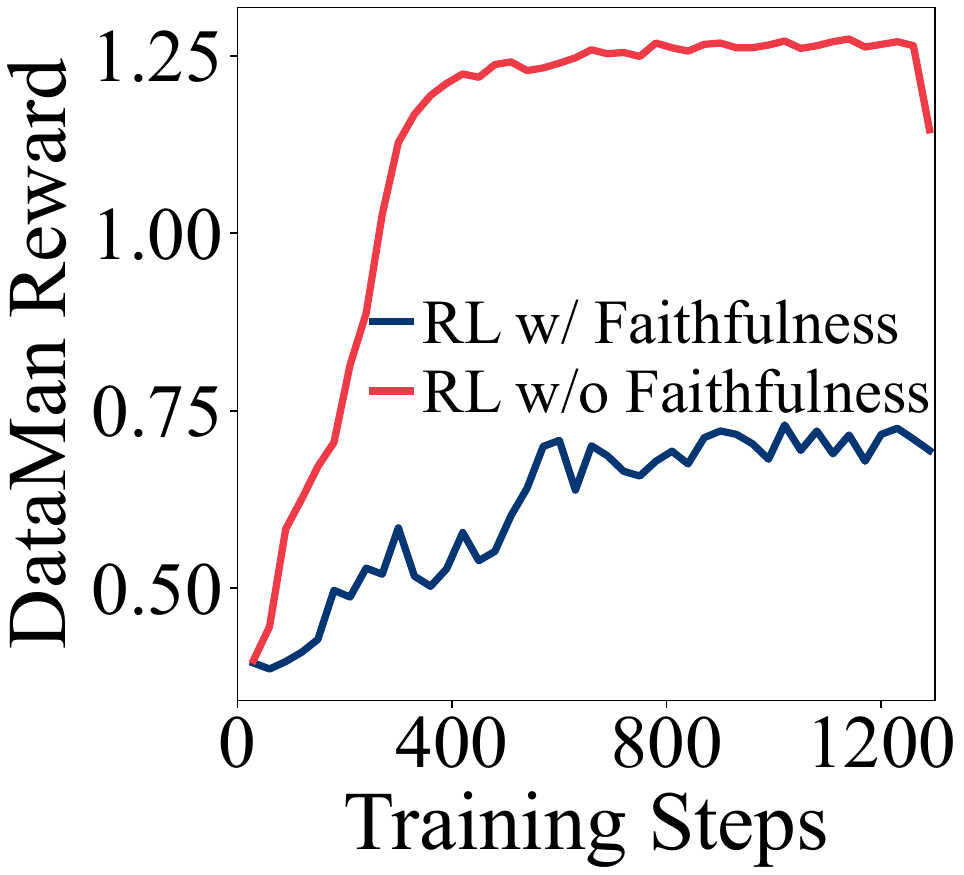}
    \caption{DataMan}
    \label{fig:rl-reward-dataman}
  \end{subfigure}
  ~
  \begin{subfigure}[t]
    {0.23\textwidth}
    \centering
    \includegraphics[width=1.0\linewidth]{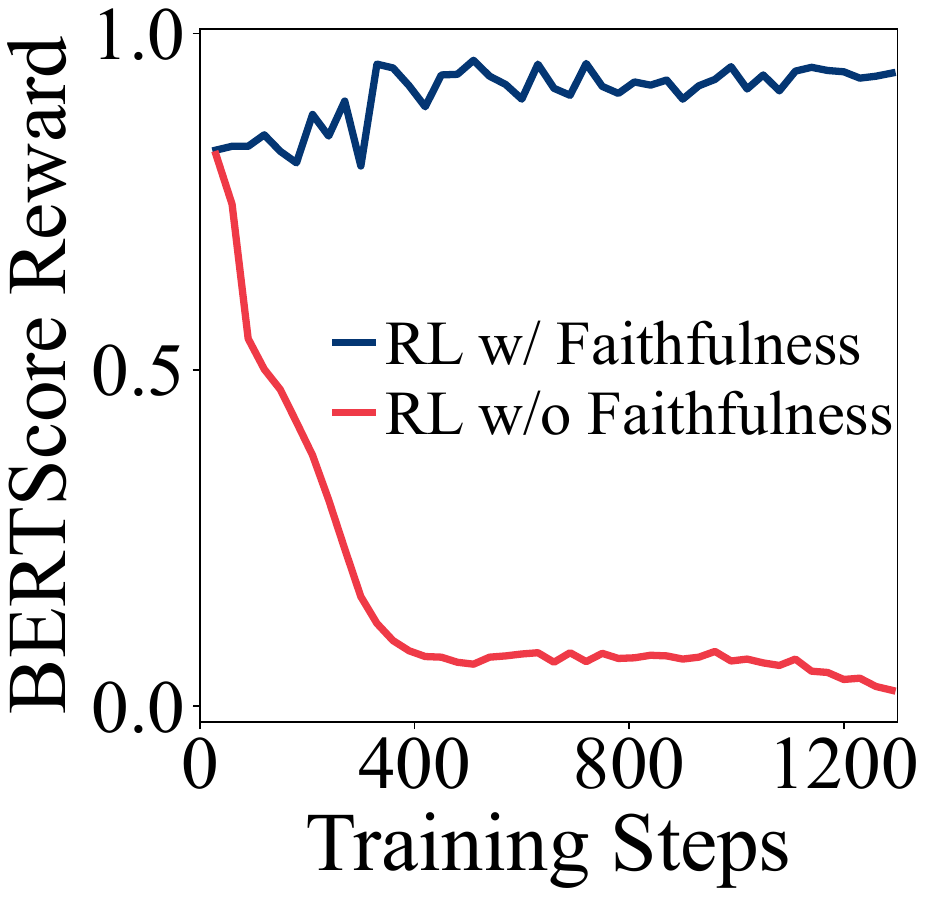}
    \caption{BERTScore}
    \label{fig:rl-reward-bertscore}
  \end{subfigure}
  ~
  \begin{subfigure}[t]
    {0.239\textwidth}
    \includegraphics[width=1.0\linewidth]{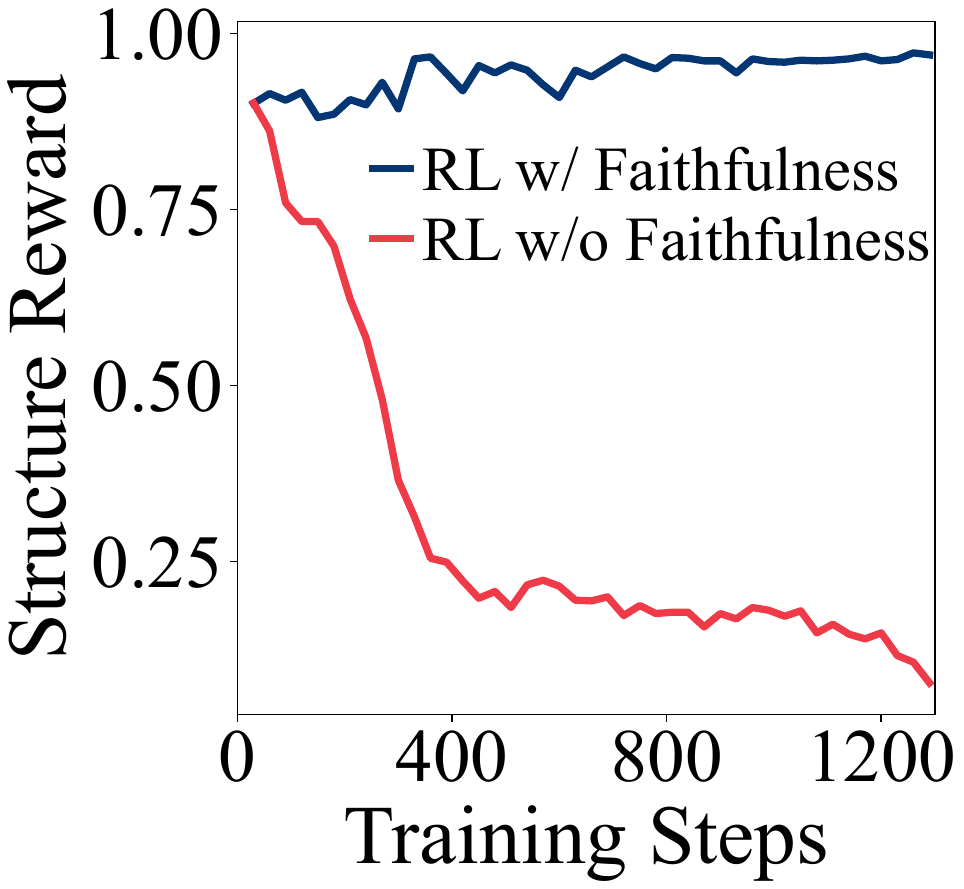}
    \caption{Structure}
    \label{fig:rl-reward-structure}
  \end{subfigure}
  ~
  \begin{subfigure}[t]
    {0.231\textwidth}
    \centering
    \includegraphics[width=1.0\linewidth]{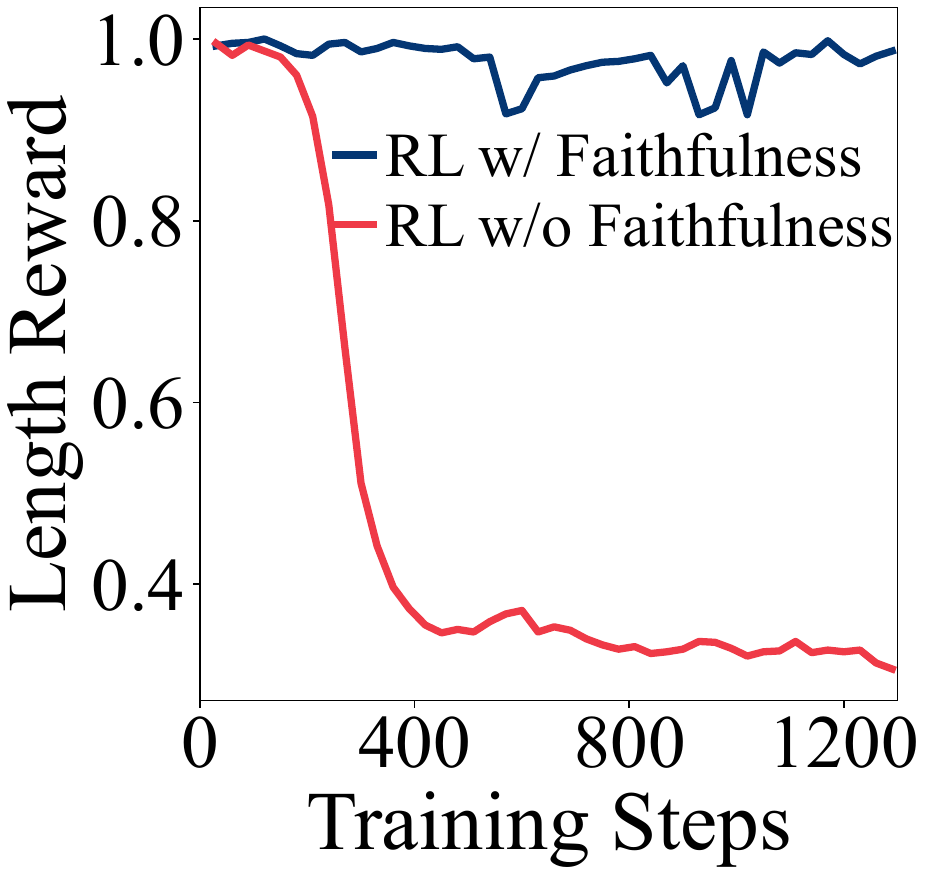}
    \caption{Length}
    \label{fig:rl-reward-len}
  \end{subfigure}
  \caption{Validation curves of (a) DataMan, (b) BERTScore, (c) structure, and (d) length
    rewards during our RL training (Qwen3-4B as base rephraser) with and without
  faithfulness rewards.}
  \label{fig:rl-dynamics}
  \vspace{-0.2cm}
\end{figure*}

This set of experiments examines the effectiveness of each reward function used in our RL
from a distributional perspective. Specifically, we plot and compare the distributions of reward-related
features for organic data and data generated by different rephrasers.
Comparisons include our base rephraser (Qwen3-4B) with direct prompting,
the rephraser trained without a specific reward,
and our final rephraser trained with all rewards.
All features are calculated on a set of 30k randomly sampled instances from our pool.

First, we compare the distributions of the DataMan score in
Figure~\ref{fig:ablation-distribution-dataman}. \textsc{RePro} effectively shifts the
distribution towards higher DataMan scores, boosting the proportion of data with a Score
of 5 from 20\% to \textbf{60\%}. In contrast, direct prompting yields a marginal
improvement over organic data, with only 25\% of recycled data achieving a Score of 5.
% This result confirms the effectiveness of our RL training in optimizing data quality.

% With faithfulness rewards, \textsc{RePro} successfully preserves the original
% characteristics at the distribution level.
\textsc{RePro} also preserves the original semantics by incorporating the BERTScore reward.
As shown in Figure~\ref{fig:ablation-distribution-bertscore}, \textsc{RePro} maintains
semantic similarity with an average BERTScore of \textbf{0.75}, compared to 0.69 for
direct prompting and 0.56 for RL without the BERTScore reward, underscoring the necessity
of this reward for semantic preservation.

We further assess structural diversity by prompting \textit{Gemini 2.5 Flash-Lite} to
classify the text structure as plain text, Markdown, blog/forum, or others. As shown in
Figure~\ref{fig:ablation-distribution-structure}, direct prompting tends to transform
Markdown-style text into plain text, whereas RL without the structure reward
over-generates Markdown-style text, harming structural diversity. The structural
distribution of our final recycled data aligns most closely with that of organic data.

Similarly, for text length (Figure~\ref{fig:ablation-distribution-len}),
we observe that direct prompting tends to produce shorter, summary-like text, while RL
without the length reward generates much longer text that may introduce extraneous
information (see Appendix~\ref{sec:case-all} for more details).
Once again, our recycled data aligns most closely with the length distribution of the
organic data, validating length faithfulness.

In Figure~\ref{fig:rl-dynamics}, we plot the training dynamics of our RL training with
and without faithfulness rewards.
The rewards shown are calculated on a validation set of 128 randomly sampled data from
$\mathcal{D}_{\text{grpo}}$, using mean value as the aggregation function.
We observe that with all rewards enabled,
DataMan, BERTScore, and structure rewards all steadily improve during training
(Figures~\ref{fig:rl-reward-dataman},~\ref{fig:rl-reward-bertscore},
and~\ref{fig:rl-reward-structure}),
while the length reward remains stably high (Figure~\ref{fig:rl-reward-len}).
This indicates that these four rewards can be effectively optimized together (instead of
trade-offs) to improve the capability of our rephraser to generate high-quality and
consistent recycled data.
In contrast, without faithfulness rewards as constraints (by setting their coefficients to zero),
DataMan reward quickly converges to a high value, but all other three rewards drop significantly, implying the overfitting issue.

In summary, these results confirm the effectiveness of each reward function in our RL framework.
They coordinate together to simultaneously improve data quality
and shape the overall distribution of recycled data to mirror that of organic data,
enabling an effective and faithful recycling.
\section{Conclusion}

In this paper, we introduce \textsc{RePro}, a novel web recycling method that trains a
small language model to generate high-quality and faithful rephrasings of web data.
Pretraining results
show that \textsc{RePro} achieves 3.7\%-14.5\% relative accuracy improvements over
organic-only baselines, doubles the gains of the state-of-the-art recycling method
ReWire, and even surpasses an organic baseline whose data pool is enlarged by 4$\times$.
Individual and distributional analyses confirm that \textsc{RePro} maintains the semantic
meaning, structural diversity, and length balance of organic data.
% Experiments with different amounts of recycled data reveal that performance peaks when
% recycling twice the organic data, effectively tripling the organic data efficiency.
% The statistics on key point recall further highlight that \textsc{RePro} preserves
% essential information more faithfully than prompting-based methods, recovering the
% highest number of supported key points and reducing omissions by up to 12$\times$
% compared to ReWire.
% Distributional analyses complement these findings, demonstrating that \textsc{RePro}
% maintains the semantic meaning, structural diversity, and length balance of organic data.
% Operation inspection shows that \textsc{RePro} enhances data quality through a combined
% use of paraphrasing, removing, and clarifying, making it versatile instead of relying
% on a predefined operation set.
% Together, these results show that \textsc{RePro} not only expands the amount of
% high-quality pretraining data but also preserves its integrity at both individual and
% distributional levels, offering a cost-efficient and reliable path to alleviate the
% data wall and enable sustainable scaling of LLM pretraining.

In conclusion, our research provides two key takeaways for effective web recycling.
First, rephrasing does not necessarily require a large language model; small models
trained with carefully designed rewards can recycle the web more effectively and
faithfully. Second, preserving the essential characteristics of organic data not only
helps maintain the original distribution but also enhances pretraining performance with
recycled data. Future work can explore more diverse and verifiable reward signals, such
as a checklist, to further improve the quality and faithfulness of recycled data, and
validate \textsc{RePro} on other web corpora such as FineWeb~\citep{penedo2024fineweb}
and RedPajama~\citep{weber2024redpajama}, which share a Common-Crawl origin and similar rule-based filtering
pipelines with DCLM-RefinedWeb.
We hope that our work inspires more cost-efficient and reliable approaches to alleviate
the data wall and enable sustainable scaling of LLM pretraining.

% \section*{Reproducibility Statement}

% We have made extensive efforts to ensure the reproducibility of our results. Anonymized
% source code is available at \url{https://anonymous.4open.science/r/RePro} to allow
% independent verification of our implementation.
% Our experimental setup is described in detail in Section~\ref{sec:setup}, including the
% dataset used, model architectures, training and inference pipelines, and
% hyperparameters in our method to facilitate faithful reproduction of our experiments.
% Additional implementation details are included in Appendix~\ref{sec:app-exp-details},
% and we reference all external resources used in our work. These materials provide the
% necessary information for others to reproduce and build upon our results.

\section*{Acknowledgments}
We thank Amazon for funding Zichun Yu through the Amazon AI Ph.D. Fellowship Program.
We thank CMU Foundation and Language Model (FLAME) Center
for providing support of computational resources.

\section*{Impact Statement}

% Authors are \textbf{required} to include a statement of the potential broader
% impact of their work, including its ethical aspects and future societal
% consequences. This statement should be in an unnumbered section at the end of
% the paper (co-located with Acknowledgements -- the two may appear in either
% order, but both must be before References), and does not count toward the paper
% page limit. In many cases, where the ethical impacts and expected societal
% implications are those that are well established when advancing the field of
% Machine Learning, substantial discussion is not required, and a simple
% statement such as the following will suffice:

% ``This paper presents work whose goal is to advance the field of Machine
% Learning. There are many potential societal consequences of our work, none
% which we feel must be specifically highlighted here.''

% The above statement can be used verbatim in such cases, but we encourage
% authors to think about whether there is content which does warrant further
% discussion, as this statement will be apparent if the paper is later flagged
% for ethics review.


This paper addresses an emerging challenge in LLM pretraining: situations in which the
available supply of high-quality organic data has been exhausted and can no longer scale
with increasing model capacity. Our work studies how scaling progress can be sustained
under such constraints by improving how organic data is effectively and efficiently
recycled, rather than prompting large models.

By training a relatively small rephraser, we reduce the computational and financial costs
of web recycling, enabling a wider range of practitioners and organizations to adopt and
further explore this approach. More broadly, the perspective of this paper encourages a
shift in pretraining practice toward methods that make data utilization both reliable and
scalable, supporting continued scaling, improving data efficiency, and enabling the
development of the next generation of frontier models under data-constrained conditions.

As with other forms of model-generated content, recycled data may reflect biases
introduced by the generating model itself. In our approach, we explicitly incorporate
faithfulness objectives into the reward design to mitigate such effects and encourage the
recycled data to closely preserve the core semantics and structure of the originals.
While this may not fully eliminate all risks associated with model-generated text, it
provides a principled mechanism to reduce unintended distortions during recycling and
helps maintain consistency between recycled and original data distributions.

Overall, our work contributes to advancing data recycling practices for pretraining and
highlights an important direction for foundation model development as high-quality data
becomes increasingly scarce.

\bibliography{bibliography}
\bibliographystyle{icml2026}

%%%%%%%%%%%%%%%%%%%%%%%%%%%%%%%%%%%%%%%%%%%%%%%%%%%%%%%%%%%%%%%%%%%%%%%%%%%%%%%
%%%%%%%%%%%%%%%%%%%%%%%%%%%%%%%%%%%%%%%%%%%%%%%%%%%%%%%%%%%%%%%%%%%%%%%%%%%%%%%
% APPENDIX
%%%%%%%%%%%%%%%%%%%%%%%%%%%%%%%%%%%%%%%%%%%%%%%%%%%%%%%%%%%%%%%%%%%%%%%%%%%%%%%
%%%%%%%%%%%%%%%%%%%%%%%%%%%%%%%%%%%%%%%%%%%%%%%%%%%%%%%%%%%%%%%%%%%%%%%%%%%%%%%
\newpage
\appendix
\onecolumn

% \chapter{}
% \addcontentsline{toc}{section}{Appendix}
% % \hypersetup{linkcolor=black}
% \etocstandarddisplaystyle
% \etocstandardlines
% \renewcommand{\contentsname}{}
% \etocsettocstyle{}{}
% \localtableofcontents

% \section{Use of Large Language Models in Paper Writing}

% \addcontentsline{toc}{section}{附录目录}
% \etocclasstocstyle
% \etocstandardlines
% \renewcommand{\contentsname}{}
% \etocsettocstyle{}{}
% \localtableofcontents

% We use LLMs to assist with writing and language refinement, specifically to improve
% grammar, phrasing, and clarity. We also use them to support our literature search and
% review. All methodological content, experiments, analysis, and interpretation were
% carried out solely by the authors. The use of LLMs was strictly limited to text
% polishing and literature review.

\makeatletter
\def\addcontentsline#1#2#3{%
  \addtocontents{#1}{%
    \protect\contentsline{#2}{#3}{\thepage}{\@currentHref}%
  }%
}
\makeatother

% \section*{Appendix}
\etocsettocstyle{\noindent\textbf{Appendix Contents}\par\medskip}{}
% \etocsettocstyle{}{}
\localtableofcontents

\section{Per-Category Discussion of Table~\ref{tab:main}}
\label{sec:app-per-category}

\textbf{For commonsense reasoning and symbolic problem solving tasks}, a coherent and
consistent logical flow within the rephrased text is preferred. This is one of
\textsc{RePro}'s strengths, as illustrated in Figure~\ref{fig:all-dataman}. ProX also
improves these tasks by removing irrelevant content, but \textsc{RePro} performs better
by flexibly applying a wider range of operations.

\textbf{For language understanding tasks}, \textsc{RePro} is notably the only web
recycling method that surpasses the organic-only baseline. We hypothesize that this is
because \textsc{RePro} is the most faithful method, so the rephrased data resembles
organic human-written text in terms of content and writing style. This aligns naturally
with the distribution of language understanding tasks, which evaluate a model's ability
to interpret human text. In contrast, other web recycling methods do not explicitly
optimize for faithfulness, leading their distributions to drift toward LM-generated text.

\textbf{For reading comprehension tasks}, rephrasings with explicit contextual
explanation may help. For instance, ReWire (which performs well on reading comprehension)
often adds phrases such as “To understand the context of the...”, which can provide
additional hints, as shown in Appendix~\ref{sec:case-all}. However, repeating similar
patterns may reduce the diversity of the training data and limit generalization.

\textbf{For world knowledge tasks}, \textsc{RePro} outperforms all other baselines but is
slightly behind ReWire. This is unsurprising because ReWire uses a much larger and
stronger model (Llama-3.3-70B-Instruct) for rephrasing, and prompting such a powerful
model can introduce additional factual knowledge from the LLM distillation. However, this
also risks reduced faithfulness, which is a known issue in prior studies. \textsc{RePro}
condenses the knowledge by removing irrelevant content and adding necessary
clarifications, as illustrated in our cases.

\section{Additional Results}
\label{sec:app-results}

This section presents additional results that complement our main findings.

% \begin{enumerate}
%     % \item Selection via DataMan.
%     \item The same Figure 3 as recycling the web for fastText.
%     \item Coefficients of different rewards.
%     % \item Different sizes of rephraser.
%     % \item Differences between rephrasing and distillation.
%     % \item Saturation of SFT.
%     % \item n-gram based diversity
%     % \item Generalize to other datasets (RedPajama-V2). I guess it needs to be put in
% the rebuttal.
% \end{enumerate}

\subsection{Efficiency Comparison}
\label{sec:app-efficiency}
\begin{wraptable}
    {r}{0.45\textwidth}
% \begin{table*}
    \centering
    \begin{minipage}{1.0\linewidth}
        \vspace{-1.1cm}
            \caption{H100 hours of rephrasing 72B tokens. We exclude ProX, as it is an operation-based method.}
            \vspace{-0.5cm}
            \label{tab:hours} \vskip 0.15in
            \resizebox{0.99\linewidth}{!}{%
            \begin{tabular}{lcccc}
                \toprule & WRAP & ReWire & \textsc{RePro}$_{\text{\textit{OLMo2-1B}}}$ & \textsc{RePro}$_{\text{\textit{Qwen3-4B}}}$ \\
                \midrule \textbf{Training} & - & - & 29 & 192 \\
                \textbf{Inference} & 2,095 & 63,360 & 698 & 1,536 \\
                \textbf{Total} & 2,095 & 63,360 & 727 & 1,728 \\
                \bottomrule
            \end{tabular}
            }
    \end{minipage}
    \vspace{-0.5cm}
\end{wraptable}
% \end{table*}

We provide a detailed comparison of the training and inference
hours required for each rephrasing method in Table~\ref{tab:hours}.
\textsc{RePro} demonstrates significant efficiency advantages over other approaches,
with up to 2.9$\times$ speedup compared to WRAP and 87.2$\times$ speedup compared to ReWire.
This confirms that our method is not only more effective but also more cost-efficient,
making it a practical choice for large-scale web recycling.

\begin{table*}[t]
    \centering
    \setlength{\tabcolsep}{3pt}
    \caption{Compute-equalized comparison in the 1B setting (1.4B model, 72B organic pool). The organic baseline selects the top-28.8B high-quality tokens from the pool and trains for 115.2B tokens, matching \textsc{RePro}'s total compute (rephrasing $+$ pretraining).}
    \vspace{-0.4cm}
    \label{tab:compute-equalized}
    \vskip 0.15in
    \centering
    \resizebox{1.0\textwidth}{!}{%
    \begin{tabular}{l|cc|cccccc}
        \toprule
                                                                                                                       &                                                                                                                                                                              &                                     & \makecell{\textbf{Commonsense}\\\textbf{Reasoning}} & \makecell{\textbf{Language}\\\textbf{Understanding}} & \makecell{\textbf{Reading}\\\textbf{Comprehension}} & \makecell{\textbf{Symbolic}\\\textbf{Problem Solving}} & \makecell{\textbf{World}\\\textbf{Knowledge}} & \makecell{\\\textbf{Core}} \\
        \textbf{Method}                                                                                                & \makecell{\textbf{Unique Data}}                                                                                                                                              & \makecell{\textbf{Training Tokens}}  & \textit{(3 tasks)}                                  & \textit{(6 tasks)}                                   & \textit{(3 tasks)}                                  & \textit{(5 tasks)}                                     & \textit{(5 tasks)}                            & \textit{(22 tasks)}        \\
        \midrule
        Organic                                                                                                        & 28.8B                                                                                                                                                                        & 115.2B                               & 0.33944                                             & 0.38274                                              & 0.21413                                             & 0.17686                                                & 0.25802                                       & 0.27871                    \\
        \rowcolor{blue!10} \textsc{RePro}$_{\text{\textit{OLMo2-1B}}}$                                                  & 7.2B + 7.2B                                                                                                                                                                  & 28.8B                                & \textbf{0.36480}                                    & \textbf{0.38379}                                     & \textbf{0.23370}                                    & \textbf{0.19162}                                       & \textbf{0.29755}                              & \textbf{0.29746}           \\
        \bottomrule
    \end{tabular}
    }
    \vspace{-0.3cm}
\end{table*}

To further verify that the gains of \textsc{RePro} are not simply a consequence of
additional compute, we equalize the total compute budget (rephrasing $+$ pretraining) and
compare against the strongest possible organic baseline. Specifically, in the 1B setting,
we select the top 28.8B high-quality tokens from the 72B organic pool and train for
115.2B tokens, which matches \textsc{RePro}'s combined rephrasing and pretraining FLOPs.
As shown in Table~\ref{tab:compute-equalized},
\textsc{RePro}$_{\text{\textit{OLMo2-1B}}}$ still outperforms this compute-equalized
organic baseline across all five task categories. This indicates that spending the budget
on rephrasing low-quality data is more beneficial than spending the same budget on
additional training over lower-quality organic data.

\subsection{Training on High-Quality Data Only}
\label{sec:app-hq-only}

\begin{table*}[t]
  \centering
  \setlength{\tabcolsep}{3pt}
  \caption{Training on high-quality organic data $\mathcal{D}_{\text{org-hq}}$ and/or
    high-quality recycled data $\mathcal{D}_{\text{rec-hq}}$ in the 400M setting.}
  \vspace{-0.4cm}
  \label{tab:hq-only}
  \vskip 0.15in
  \centering
  \resizebox{1.0\textwidth}{!}{%
    \begin{tabular}{l|cc|cccccc}
      \toprule
      &
      &
      & \makecell{\textbf{Commonsense}\\\textbf{Reasoning}} &
      \makecell{\textbf{Language}\\\textbf{Understanding}} &
      \makecell{\textbf{Reading}\\\textbf{Comprehension}} &
      \makecell{\textbf{Symbolic}\\\textbf{Problem Solving}} &
      \makecell{\textbf{World}\\\textbf{Knowledge}} & \makecell{\\\textbf{Core}} \\
      \textbf{Training Data}
      & \makecell{\textbf{Pool}}
      & \makecell{\textbf{Unique Data}}
      & \textit{(3 tasks)}                                  & \textit{(6 tasks)}
      & \textit{(3 tasks)}                                  & \textit{(5 tasks)}
      & \textit{(5 tasks)}                            & \textit{(22 tasks)}        \\
      \midrule
      $\mathcal{D}_{\text{org-hq}}$
      & 72B
      & 7.2B
      & 0.23613                                             & 0.27079
      & 0.03724                                             & 0.14535
      & 0.20126                                       & 0.18990                    \\
      $\mathcal{D}_{\text{rec-hq}}$ (ReWire)
      & 72B
      & 7.2B
      & 0.23824                                             & 0.22793
      & 0.05802                                             & 0.15852
      & 0.21445                                       & 0.18733                    \\
      $\mathcal{D}_{\text{rec-hq}}$ (\textsc{RePro}$_{\text{\textit{OLMo2-1B}}}$)
      & 72B
      & 7.2B
      & 0.25099                                             & 0.26855
      & 0.07980                                             & 0.15949
      & 0.20993                                       & 0.20231                    \\
      $\mathcal{D}_{\text{org-hq}}$ $+$ $\mathcal{D}_{\text{rec-hq}}$ (ReWire)
      & 72B
      & 7.2B $+$ 7.2B
      & 0.24051                                             & 0.26453
      & 0.06232                                             & 0.17392
      & 0.21246                                       & 0.20125                    \\
      \rowcolor{blue!10} $\mathcal{D}_{\text{org-hq}}$ $+$ $\mathcal{D}_{\text{rec-hq}}$
      (\textsc{RePro}$_{\text{\textit{OLMo2-1B}}}$) & 72B
      & 7.2B $+$ 7.2B
      & \textbf{0.27361}                                    & \textbf{0.27692}
      & \textbf{0.10292}                                    & \textbf{0.17819}
      & \textbf{0.22057}                              & \textbf{0.21749}           \\
      \bottomrule
    \end{tabular}
  }
  \vspace{-0.3cm}
\end{table*}

A natural question is whether \textsc{RePro}'s gains come purely from the rephrased data,
in which case one could discard the organic data entirely. To evaluate this, we train 400M
models on $\mathcal{D}_{\text{org-hq}}$ alone, on $\mathcal{D}_{\text{rec-hq}}$ alone
(for both ReWire and \textsc{RePro}), and on the two merged mixtures
$\mathcal{D}_{\text{org-hq}} \cup \mathcal{D}_{\text{rec-hq}}$ (our primary setting). As shown in
Table~\ref{tab:hq-only}, $\mathcal{D}_{\text{rec-hq}}$ from \textsc{RePro} alone already
outperforms $\mathcal{D}_{\text{org-hq}}$ alone by +6.5\% on Core, whereas
$\mathcal{D}_{\text{rec-hq}}$ from ReWire alone slightly underperforms
$\mathcal{D}_{\text{org-hq}}$ due to its limited faithfulness. However, the best result is still obtained by
\emph{merging} $\mathcal{D}_{\text{org-hq}}$ with $\mathcal{D}_{\text{rec-hq}}$ from
\textsc{RePro}, indicating that recycled data acts as a \emph{complement} to organic data rather than a replacement. In practice, since the bottleneck for frontier
LLMs is limited high-quality organic data, merging both sources is the realistic way to
mitigate data scarcity.

\subsection{Controlled Comparison with ReWire}
\label{sec:app-controlled-rewire}

\begin{table*}[t]
    \centering
    \setlength{\tabcolsep}{3pt}
    \caption{Controlled comparison between ReWire and \textsc{RePro} in the 400M setting (400M model, 72B organic pool). ``Same'' uses the same 72B organic pool for both methods' rephrasing; ``Separate'' lets each method rephrase its own independent 72B pool, so the union effectively doubles the organic data pool, matching the advantage ReWire had in our main results.}
    \vspace{-0.4cm}
    \label{tab:controlled-rewire}
    \vskip 0.15in
    \centering
    \resizebox{1.0\textwidth}{!}{%
    \begin{tabular}{l|cc|cccccc}
        \toprule
                                                                                                                       &                                                                                                                                                                              &                                                                                                                                                                              & \makecell{\textbf{Commonsense}\\\textbf{Reasoning}} & \makecell{\textbf{Language}\\\textbf{Understanding}} & \makecell{\textbf{Reading}\\\textbf{Comprehension}} & \makecell{\textbf{Symbolic}\\\textbf{Problem Solving}} & \makecell{\textbf{World}\\\textbf{Knowledge}} & \makecell{\\\textbf{Core}} \\
        \textbf{Method}                                                                                                & \makecell{\textbf{Rephraser}}                                                                                                                                                & \makecell{\textbf{Rephrasing Pool}}                                                                                                                                          & \textit{(3 tasks)}                                  & \textit{(6 tasks)}                                   & \textit{(3 tasks)}                                  & \textit{(5 tasks)}                                     & \textit{(5 tasks)}                            & \textit{(22 tasks)}        \\
        \midrule
        ReWire                                                                                                         & Qwen3-4B                                                                                                                                                                     & Same                                                                                                                                                                         & 0.24105                                             & 0.26162                                              & 0.05101                                             & 0.17632                                                & 0.20887                                       & 0.19872                    \\
        \rowcolor{blue!10} \textsc{RePro}                                                                              & Tuned Qwen3-4B                                                                                                                                                               & Same                                                                                                                                                                         & \textbf{0.28454}                                    & \textbf{0.27792}                                     & \textbf{0.07181}                                    & \textbf{0.19409}                                       & \textbf{0.21154}                              & \textbf{0.21658}           \\
        \midrule
        ReWire                                                                                                         & Llama-3.3-70B-Instruct                                                                                                                                                       & Separate                                                                                                                                                                     & 0.24051                                             & 0.26453                                              & 0.06232                                             & 0.17392                                                & 0.21246                                       & 0.20125                    \\
        \rowcolor{blue!10} \textsc{RePro}                                                                              & Tuned Qwen3-4B                                                                                                                                                               & Separate                                                                                                                                                                     & \textbf{0.29121}                                    & \textbf{0.27634}                                     & \textbf{0.09488}                                    & \textbf{0.18784}                                       & \textbf{0.21825}                              & \textbf{0.22031}           \\
        \bottomrule
    \end{tabular}
    }
    \vspace{-0.3cm}
\end{table*}

In our main results (Table~\ref{tab:main}), ReWire uses its publicly released recycled
data, which was rephrased from a different organic pool than ours. This gives ReWire an
implicit advantage, since its method effectively benefits from a larger combined organic
pool. To remove this confound, we run two controlled comparisons in the 400M setting,
summarized in Table~\ref{tab:controlled-rewire}. First, we apply ReWire on \emph{the
same} 72B organic pool used by \textsc{RePro}, with Qwen3-4B (with thinking) as the
rephraser to match \textsc{RePro}'s base model. \textsc{RePro} outperforms ReWire under
this matched setup by 9.0\% on Core. Second, we let each method rephrase its own
independent 72B organic pool (so the union of organic pools doubles in size, mirroring
the advantage ReWire had in our main results); \textsc{RePro} still outperforms ReWire
(Llama-3.3-70B-Instruct) across all categories. Together, these results indicate that
\textsc{RePro}'s gains over ReWire are not an artifact of differing organic-data sources.

\subsection{Significance Analysis}
\label{sec:app-variance}

\begin{table*}[t]
    \centering
    \setlength{\tabcolsep}{3pt}
    \caption{Variance of the organic baseline across three random seeds in the 400M setting (400M model, 28.8B training tokens), reported alongside \textsc{RePro}$_{\text{\textit{OLMo2-1B}}}$ for reference. The bottom row gives the standard deviation of the organic baseline.}
    \vspace{-0.4cm}
    \label{tab:variance}
    \vskip 0.15in
    \centering
    \resizebox{1.0\textwidth}{!}{%
    \begin{tabular}{l|ccc|cccccc}
        \toprule
                                                                                                                       &                                                                                                                                                                              &                                  &                                                                                                                                                                              & \makecell{\textbf{Commonsense}\\\textbf{Reasoning}} & \makecell{\textbf{Language}\\\textbf{Understanding}} & \makecell{\textbf{Reading}\\\textbf{Comprehension}} & \makecell{\textbf{Symbolic}\\\textbf{Problem Solving}} & \makecell{\textbf{World}\\\textbf{Knowledge}} & \makecell{\\\textbf{Core}} \\
        \textbf{Method}                                                                                                & \makecell{\textbf{Pool}}                                                                                                                                                     & \makecell{\textbf{Unique Data}}  & \makecell{\textbf{Seed}}                                                                                                                                                     & \textit{(3 tasks)}                                  & \textit{(6 tasks)}                                   & \textit{(3 tasks)}                                  & \textit{(5 tasks)}                                     & \textit{(5 tasks)}                            & \textit{(22 tasks)}        \\
        \midrule
        Organic                                                                                                        & 72B                                                                                                                                                                          & 7.2B                             & 13                                                                                                                                                                           & 0.23738                                             & 0.26935                                              & 0.04440                                             & 0.14056                                                & 0.20129                                       & 0.18957                    \\
        Organic                                                                                                        & 72B                                                                                                                                                                          & 7.2B                             & 87                                                                                                                                                                           & 0.23827                                             & 0.27080                                              & 0.03554                                             & 0.14490                                                & 0.20212                                       & 0.19006                    \\
        Organic                                                                                                        & 72B                                                                                                                                                                          & 7.2B                             & 42                                                                                                                                                                           & 0.23613                                             & 0.27079                                              & 0.03724                                             & 0.14535                                                & 0.20126                                       & 0.18990                    \\
        \rowcolor{blue!10} \textsc{RePro}$_{\text{\textit{OLMo2-1B}}}$                                                  & 72B                                                                                                                                                                          & 7.2B + 7.2B                      & 42                                                                                                                                                                           & \textbf{0.27361}                                    & \textbf{0.27692}                                     & \textbf{0.10292}                                    & \textbf{0.17819}                                       & \textbf{0.22057}                              & \textbf{0.21749}           \\
        \midrule
        Organic (std)                                                                                                  & --                                                                                                                                                                           & --                               & --                                                                                                                                                                           & $\pm$1.08e-3                                        & $\pm$8.34e-4                                         & $\pm$4.70e-3                                        & $\pm$2.65e-3                                           & $\pm$4.88e-4                                  & $\pm$2.50e-4               \\
        \bottomrule
    \end{tabular}
    }
    \vspace{-0.3cm}
\end{table*}

To quantify the noise floor of our main results, we resample the organic data pool with
three different random seeds in the 400M setting and report the per-category accuracy and
standard deviation in Table~\ref{tab:variance}. \textsc{RePro}'s gains over the organic
baseline exceed twice the standard deviation in every category, including the
higher-variance ones (Reading Comprehension, Symbolic Problem Solving), and in the
aggregate Core score. Due to compute and time constraints, we are unable to rerun the
full recycling pipeline across multiple seeds; the organic-side variance therefore serves
as a conservative noise floor.

\subsection{Improvements on All DataMan Criteria}

\begin{figure*}[t]
  \centering
  % \vspace{-0.2cm}
  \includegraphics[width=1.0\linewidth]{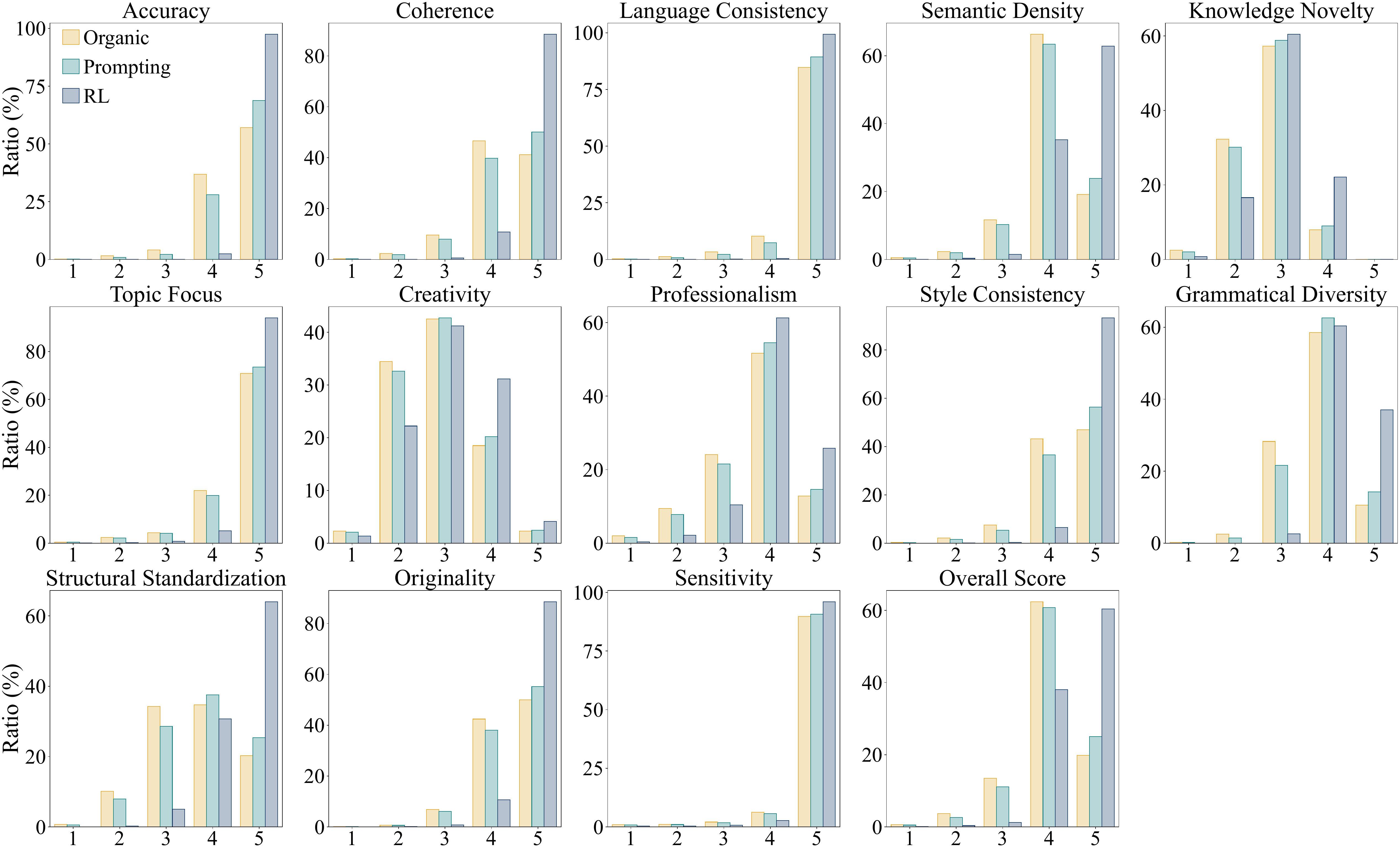}
  \vspace{-0.5cm}
  \caption{Improvements of \textsc{RePro}$_{\text{\textit{Qwen3-4B}}}$ on all DataMan criteria.}
  \label{fig:all-dataman}
  \vspace{-0.3cm}
\end{figure*}

In this section, we present the improvements of \textsc{RePro} on all DataMan criteria.
As shown in Figure~\ref{fig:all-dataman},
despite only being optimized for the overall score, \textsc{RePro} consistently
outperforms the organic data and prompting baseline across all individual criteria.
For some subjective criteria like \textit{knowledge novelty} and \textit{creativity},
the ratio of Score=5 remains low after rephrasing.
This is expected, as these aspects are inherently challenging to enhance through rephrasing alone,
especially given our rephraser is trained to faithfully preserve the original content.
In summary, these results highlight the effectiveness of \textsc{RePro} in improving
various dimensions of data quality.

\subsection{Hyperparameter Studies}
\label{sec:app-hyper}

We determine the reward thresholds and coefficients in two steps. First, we initialize
them from distributional observations on prompting-based rephrasings. Second, we refine
them by monitoring validation reward curves during RL, ensuring that all four rewards can
be jointly optimized rather than collapsing onto the DataMan signal. This is a one-time
overhead common to RL-based methods, and we share all final values to facilitate future research.

\begin{figure*}[t]
  \centering
  % \vspace{-0.3cm}
  \begin{subfigure}[t]
    {0.234\textwidth}
    \centering
    \includegraphics[width=1.0\linewidth]{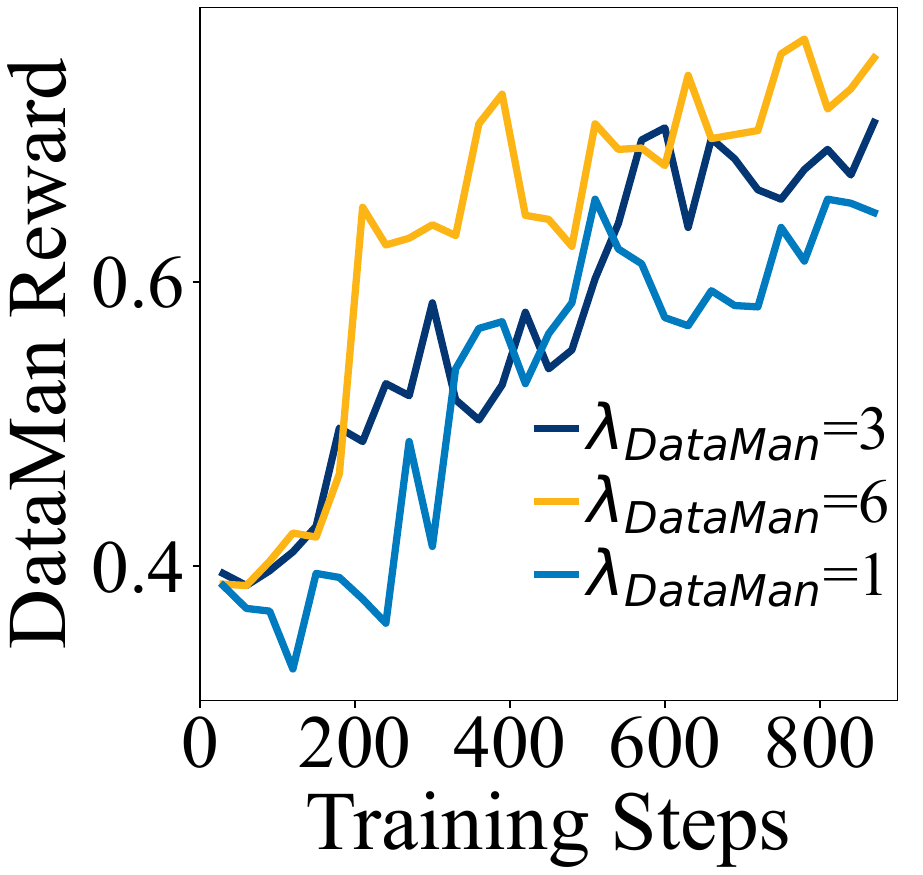}
    \caption{DataMan}
  \end{subfigure}
  ~
  \begin{subfigure}[t]
    {0.228\textwidth}
    \centering
    \includegraphics[width=1.0\linewidth]{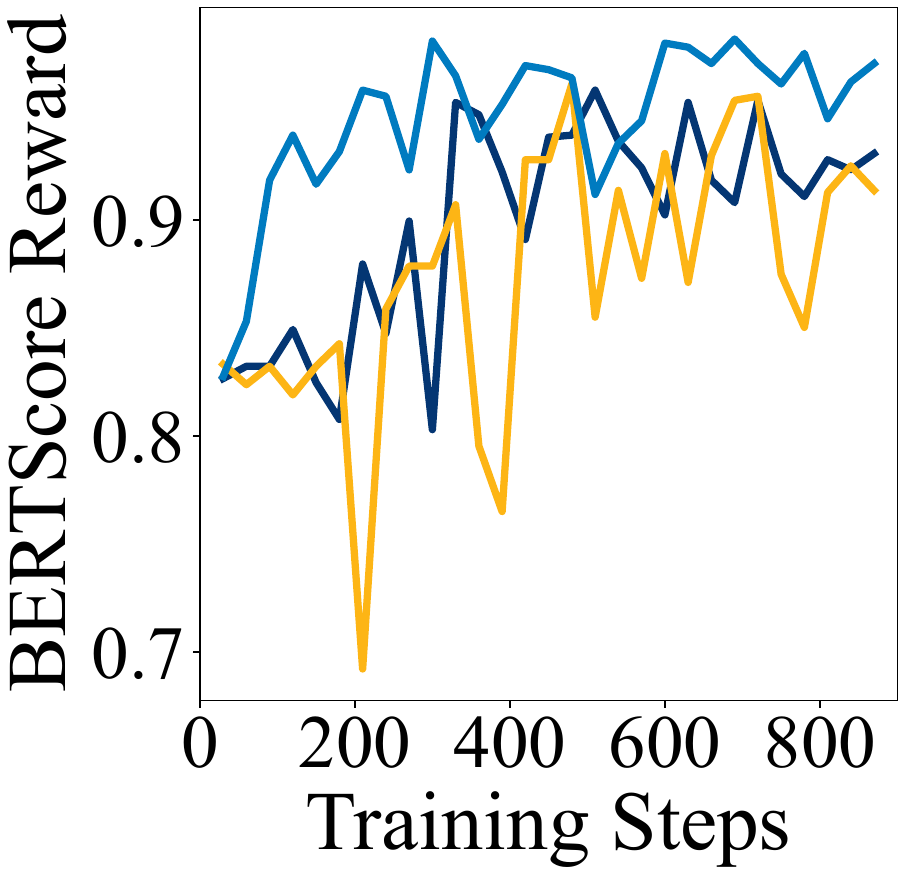}
    \caption{BERTScore}
  \end{subfigure}
  ~
  \begin{subfigure}[t]
    {0.237\textwidth}
    \includegraphics[width=1.0\linewidth]{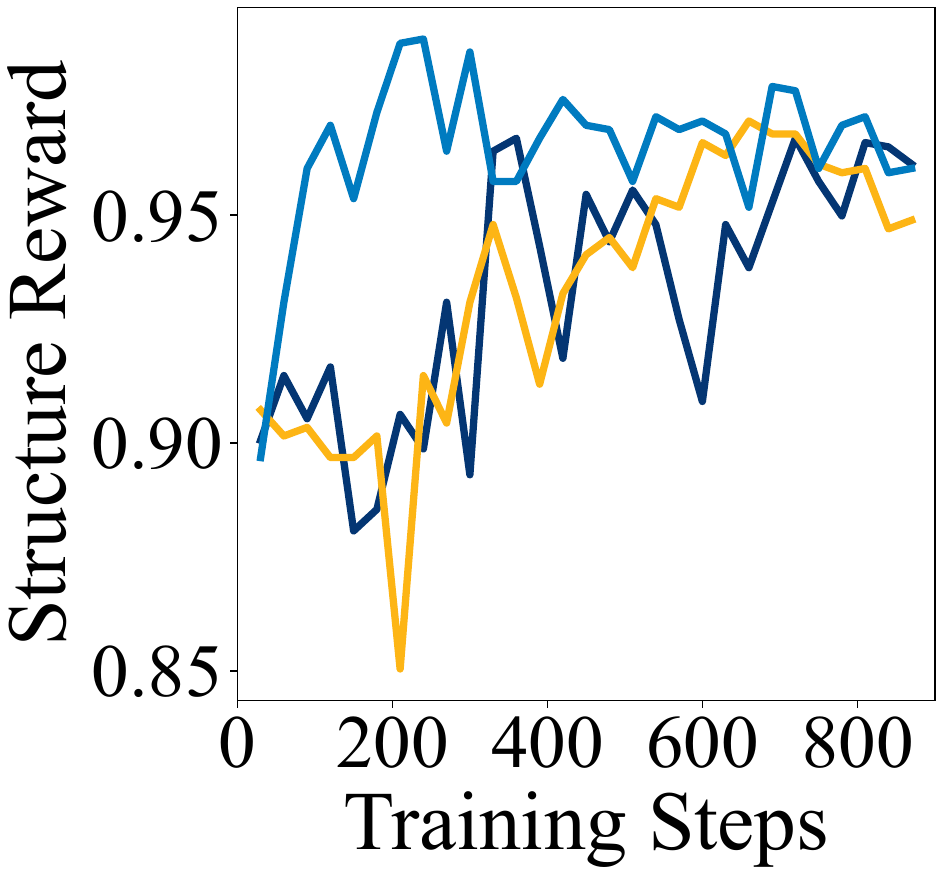}
    \caption{Structure}
  \end{subfigure}
  ~
  \begin{subfigure}[t]
    {0.246\textwidth}
    \centering
    \includegraphics[width=1.0\linewidth]{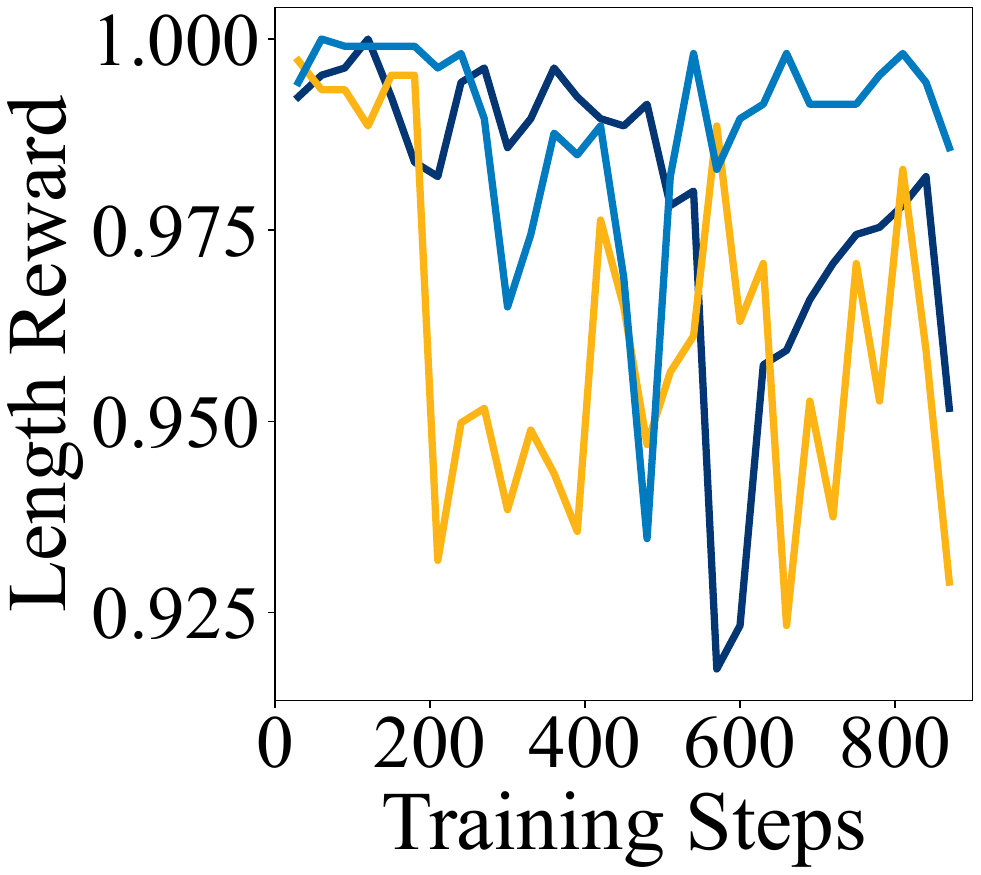}
    \caption{Length}
  \end{subfigure}
  \caption{Validation curves of (a) DataMan, (b) BERTScore, (c) structure, and (d) length
  rewards during our RL training (Qwen3-4B as base rephraser) with different DataMan coefficients.}
  \label{fig:rl-dynamics-ablation}
  \vspace{-0.2cm}
\end{figure*}

As a representative study, we vary the coefficient of DataMan reward
$\lambda_{\text{DataMan}}$ to 1, 3 (our main setup), and 6.
As shown in Figure~\ref{fig:rl-dynamics-ablation}, all reward curves increase steadily
and show a very similar pattern, showing the robustness of our method. Upweighting
DataMan reward can make it increase faster, yet at the cost of larger fluctuations in
faithfulness rewards, and vice versa.
Our final choice strikes a decent balance between DataMan and faithfulness rewards.
Our compute resources do not support a larger RL, e.g., training Qwen3-8B, while
concurrent work~\citep{maini2025beyondweb} finds a diminishing return when using an 8B
rephraser compared to 3B.

For $\tau_{\text{BERTScore}}$, we set it as 0.65, since we find that about two-thirds of
the recycled data by prompting can achieve a BERTScore above this value. It is a
representative cutoff that reflects the majority semantic similarity level of the
recycled data while still leaving room to improve rephrasing quality.

To better understand the effect of sampling, we vary the temperature to 0.6 and the top-p
value to 0.96 for the prompting baseline. A temperature of 0.6 produces more
deterministic and consistent outputs, while top-p 0.96 leads to more diverse and creative
sampling. As shown in Table~\ref{tab:sampling}, top-p=0.96 performs slightly better than
the other two, likely due to the increased diversity in the recycled data, although
overall differences are small. \textsc{RePro} remains clearly stronger than all
prompting-based variants, showing that RL is a more practical path for training an
effective rephraser.

\begin{table*}
    [t]
    \setlength{\tabcolsep}{3pt}
    \caption{Ablation studies on different sampling hyperparameters in the 400M setup (Qwen3-4B as base rephraser).}
    \vspace{-0.6cm}
    \label{tab:sampling} 
    \vskip 0.15in
    % \begin{sc}
        \centering
        \resizebox{1.0\textwidth}{!}{%
        \begin{tabular}{l|cc|cccccc}
            \toprule                                    & & & \makecell{\textbf{Commonsense}\\\textbf{Reasoning}} & \makecell{\textbf{Language}\\\textbf{Understanding}} & \makecell{\textbf{Reading}\\\textbf{Comprehension}} & \makecell{\textbf{Symbolic}\\\textbf{Problem Solving}} & \makecell{\textbf{World}\\\textbf{Knowledge}} & \makecell{\\\textbf{Core}} \\
            \textbf{Method} &
            \makecell{\textbf{Pool}} & \makecell{\textbf{Unique Data}} & \textit{(3 tasks)}                                  & \textit{(6 tasks)}                                   & \textit{(3 tasks)}                                  & \textit{(5 tasks)}                                     & \textit{(5 tasks)}                            & \textit{(22 tasks)}        \\
            \midrule                                                        
            \multicolumn{6}{l}{\textbf{400M Setting:} 400M model, 28.8B training tokens}        \\
            \midrule
            Organic                                                                       & 72B & 14.4B & 0.26953 & 0.25781 & 0.05623 & 0.14991 & 0.17683 & 0.18899
            \\             
             
            Prompting (temperature=1.0, top-p=0.9)                                               & 72B & 7.2B + 7.2B & 0.24310 & 0.26758 & 0.05075 & 0.17392 & 0.20196 & 0.19847
            \\           
             
            Prompting (temperature=0.6, top-p=0.9)                                               & 72B & 7.2B + 7.2B & 0.26988 & 0.25236 & 0.07081 & 0.16191 & 0.19956 & 0.19743
            \\             
             
            Prompting (temperature=1.0, top-p=0.96)                                               & 72B & 7.2B + 7.2B & 0.26594 & 0.27233 & 0.04183 & 0.16991 & 0.20038 & 0.20040
            \\
           
            \textsc{RePro}                                               & 72B & 7.2B + 7.2B & \textbf{0.28454} & \textbf{0.27792} & \textbf{0.07181} & \textbf{0.19409} & \textbf{0.21154} & \textbf{0.21658}
            \\
            \bottomrule
        \end{tabular}
        }
    % \end{sc}
\end{table*}

\subsection{Choice of Quality Reward}
\begin{table*}
    [ht]
    \setlength{\tabcolsep}{3pt}
    \caption{Data selection with different quality functions.}
    \vspace{-0.6cm}
    \label{tab:dataman} 
    \vskip 0.15in
    % \begin{sc}
        \centering
        \resizebox{1.0\textwidth}{!}{%
        \begin{tabular}{l|cc|cccccc}
            \toprule                                    & & & \makecell{\textbf{Commonsense}\\\textbf{Reasoning}} & \makecell{\textbf{Language}\\\textbf{Understanding}} & \makecell{\textbf{Reading}\\\textbf{Comprehension}} & \makecell{\textbf{Symbolic}\\\textbf{Problem Solving}} & \makecell{\textbf{World}\\\textbf{Knowledge}} & \makecell{\\\textbf{Core}} \\
            \textbf{Quality Function} &
            \makecell{\textbf{Pool}} & \makecell{\textbf{Unique Data}} & \textit{(3 tasks)}                                  & \textit{(6 tasks)}                                   & \textit{(3 tasks)}                                  & \textit{(5 tasks)}                                     & \textit{(5 tasks)}                            & \textit{(22 tasks)}        \\
            \midrule                                                        
            \multicolumn{6}{l}{\textbf{1B Setting:} 1.4B model, 28.8B training tokens}        \\
            \midrule
            Random                                                                       & 72B & 7.2B & 0.28256                                             & 0.33930                                              & 0.19857                                             & 0.16795                                                & 0.23611                                       & 0.24998                    \\
            DCLM-fastText                                                                       & 72B & 7.2B & 0.32348                                             & \textbf{0.38371}                                              & 0.19584                                             & \textbf{0.19795}                                                & \textbf{0.28745}                                       & \textbf{0.28578}                    \\
            DCLM-fastText                                                                       & 72B & 14.4B & 0.31483                                             & 0.37916                                              & 0.15112                                             & 0.17507                                                & 0.28311                                       & 0.27108 \\
            DataMan                                                                       & 72B & 7.2B & 0.31369                                             & 0.32591                                              & 0.17524                                             & 0.16676                                                & 0.27709                                       & 0.25643 \\
            DCLM-fastText + DataMan                                                                       & 72B & 7.2B + 7.2B & \textbf{0.32643}                                             & 0.37210                                              & \textbf{0.19902}                                             & 0.19010                                                & 0.28630                                       & 0.28141 \\
            \bottomrule
        \end{tabular}
        }
    % \end{sc}
\end{table*}

In this section, we explore the impact of using other quality rewards in our RL training.
Specifically, we investigate two additional quality reward options:
(1) DCLM-fastText, a n-gram-based classifier that performs the best in data selection
from DCLM-RefinedWeb,
and (2) training data influence~\citep{yu2024mates}, which measures the actual training
effect of each data point on the reference task given a pretrained model checkpoint.
Following~\citet{yu2025group}, we use our 400M organic baseline at 10k steps as the model
checkpoint and adopt FLAN~\citep{wei2022flan} as the reference task.

\begin{wrapfigure}
  {r}{0.3\textwidth}
  \centering
  \vspace{-0.6cm}
  \includegraphics[width=1.0\linewidth]{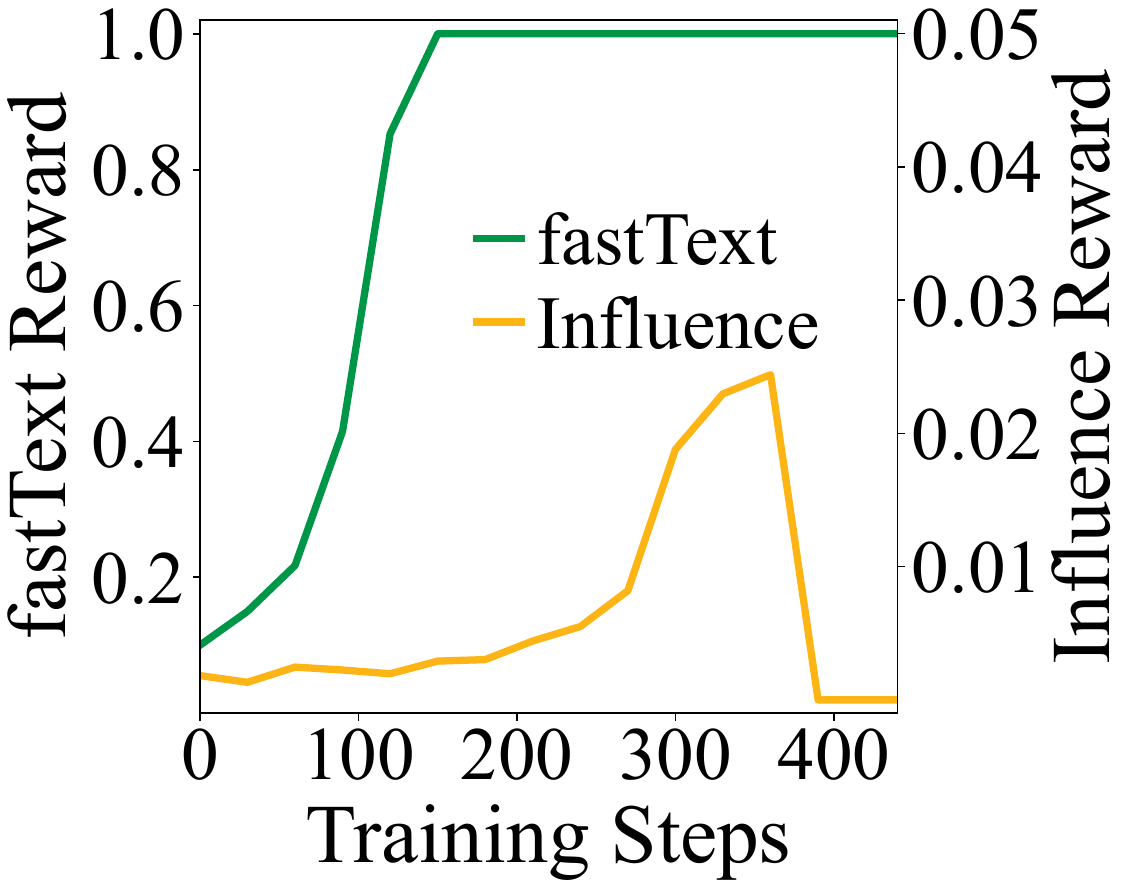}
  \vspace{-0.6cm}
  \caption{RL (Qwen3-4B as base rephraser) with DCLM-fastText or data influence as quality rewards.}
  \label{fig:quality}
  \vspace{-0.4cm}
\end{wrapfigure}

As illustrated in Figure~\ref{fig:quality}, the fastText reward quickly saturates at a high value,
while the influence reward remains low and drops sharply to 0 (no rephrasing) at the end.
A case study in Appendix~\ref{sec:case-all} shows that the rephraser trained with
fastText reward tends to generate text with an academic tone,
which hacks the n-gram-based classifier to get high scores.
In contrast, the rephraser trained with the influence reward struggles to explore
rephrasing strategies that consistently improve the reward, likely because this signal is
too fine-grained and difficult to optimize upon.
Our findings suggest that a useful metric for data selection may not be an appropriate
reward for rephraser RL.
A good quality reward should not only correlate well with the final pretraining performance,
but also be robust to reward hacking and can be effectively optimized.

We also evaluate the effectiveness of using DataMan as a quality function for selecting
organic data.
As shown in Table~\ref{tab:dataman},
either using DataMan selection only or combining it with
DCLM-fastText selection could not beat using DCLM-fastText alone.
This suggests that (1)
simply merging multiple quality selections may not enlarge the amount of high-quality data, and
(2) the reward in rephraser training is not necessarily the best quality function for
data selection.

% \input{tables/operations}

% \input{tables/operations}

% An interesting case study in Appendix~\ref{sec:app-cases} further illustrates this phenomenon,
% where the recycled text without faithfulness rewards introduces extraneous information
% loosely related to the original text and alters the structure of the the organic data
% from markdown to plain text. This may increase the DataMan score but harms the overall
% faithfulness of recycled data.This may increase the DataMan score but harms the overall
% faithfulness of recycled data.
% In contrast, \textsc{RePro} strictly adheres to the original content and structure
% while improving the writing quality by removing irrelevant advertisements and metadata.

% tends to generate text that is only loosely related to the original text but has a decent writing,
% which increases the DataMan score but harms the overall faithfulness of recycled data.

% \input{tables/unique}

\section{Additional Experiment Details}
\label{sec:app-exp-details}

This section provides additional configuration details about our experiments.

\subsection{Training Hyperparameters}

We provide our training hyperparameters in Table~\ref{tab:config}.

\begin{table}[ht]
    \centering
    \begin{minipage}{0.7\linewidth}
            \caption{Training details.}
            \vspace{-0.6cm}
            \label{tab:config} \vskip 0.15in
            \resizebox{0.99\linewidth}{!}{%
            \begin{tabular}{l|cccc}
                \toprule \textbf{Hyperparameter} & \textbf{400M LM} & \textbf{1.4B LM} & \textbf{2.8B LM} & \textbf{Rephraser} \\
                \midrule Steps            & 27462            & 54923       & 106604         & 2000                                      \\
                Batch size                & 512              & 256         & 256         & 24                                        \\
                Sequence length           & 2048             & 2048        & 2048         & 4096 (2048i + 2048o)                                                                \\
                Max learning rate         & 3e-3             & 3e-3        & 3e-3         & 1e-6                                      \\
                Optimizer                 & AdamW            & AdamW       & AdamW         & AdamW                                     \\
                Scheduler                 & Cosine           & Cosine      & Cosine         & Cosine  
                                            \\
                \bottomrule
            \end{tabular}
            }
    \end{minipage}
\end{table}

\subsection{Evaluation Tasks}
\label{sec:app-tasks}

We list all evaluation task names and their few-shot numbers in Table~\ref{tab:tasks}.

\begin{table}
    [ht]
    \setlength{\tabcolsep}{3pt}
    \caption{All evaluation task names and their few-shot numbers.}
    \vspace{-0.6cm}
    \label{tab:tasks} 
    \vskip 0.15in
    % \begin{sc}
        \centering
    \begin{tabular}{llc}
    \toprule
    \textbf{Category} & \textbf{Task} & \textbf{\#Shots} \\
    \midrule
    
    \multirow{3}{*}{Commonsense Reasoning} 
    & copa~\citep{roemmele2011choice} & 0 \\
    & openbook\_qa~\citep{OpenBookQA2018} & 0 \\
    & piqa~\citep{bisk2020piqa} & 10 \\
    
    \midrule
    \multirow{6}{*}{Language Understanding} 
    & bigbench\_language\_identification~\citep{srivastava2023bbh} & 10 \\
    & hellaswag\_zeroshot~\citep{zellers2019hellaswag} & 0 \\
    & hellaswag~\citep{zellers2019hellaswag} & 10 \\
    & lambada\_openai~\citep{paperno2016lambada} & 0 \\
    & winograd~\citep{levesque2012winograd} & 0 \\
    & winogrande~\citep{sakaguchi2021winogrande} & 0 \\
    
    \midrule
    \multirow{3}{*}{Reading Comprehension} 
    & boolq~\citep{clark2019boolq} & 10 \\
    & coqa~\citep{reddy2019coqa} & 0 \\
    & squad~\citep{rajpurkar2016squad} & 10 \\
    
    \midrule
    \multirow{5}{*}{Symbolic Problem Solving} 
    & agi\_eval\_lsat\_ar~\citep{zhong2024agieval} & 3 \\
    & bigbench\_cs\_algorithms~\citep{srivastava2023bbh} & 10 \\
    & bigbench\_dyck\_languages~\citep{srivastava2023bbh} & 10 \\
    & bigbench\_operators~\citep{srivastava2023bbh} & 10 \\
    & bigbench\_repeat\_copy\_logic~\citep{srivastava2023bbh} & 10 \\
    
    \midrule
    \multirow{5}{*}{World Knowledge} 
    & arc\_challenge~\citep{clark2018arc} & 10 \\
    & arc\_easy~\citep{clark2018arc} & 10 \\
    & bigbench\_qa\_wikidata~\citep{srivastava2023bbh} & 10 \\
    & jeopardy~\citep{kaggle2019jeopardy} & 10 \\
    & mmlu\_fewshot~\citep{hendryckstest2021} & 5 \\
    
    \bottomrule
    \end{tabular}
\end{table}

\subsection{Baselines}
\label{sec:app-baselines}

We provide the implementation details of the baselines we compare against in our main results:

\begin{itemize}
  \item WRAP~\citep{maini2024rephrasing}: Following their original paper, we adopt
    Mistral-7B-Instruct-v0.1~\citep{jiang2023mistral7b} as the rephraser, using their
    Wikipedia-style prompt.
  \item ProX~\citep{zhou2024prox}: This method uses
    Llama-3-70B-Instruct~\citep{dubey2024llama} to annotate chunk-level programs to
    polish the organic data. Then, they train a 0.3B model to learn the annotated
    programs and perform final edits. We emphasize that their recycling mechanism relies
    on program-based operations, in contrast to our generation-based rephrasing. These
    two mechanisms are essentially orthogonal, and we believe future work can leverage
    both to more effectively recycle the web.
  \item ReWire~\citep{nguyen2025rewire}: This method employs guided rewriting with
    chain-of-thought reasoning using Llama-3.3-70B-Instruct~\citep{dubey2024llama}. As
    their code has not been open-sourced, we randomly sample 7.2B tokens from their
    released data. This may give them a performance advantage, since their organic pool
    differs from ours, meaning their method in our main results actually utilizes a larger pool.
\end{itemize}

\newpage

\subsection{Prompts}
\label{sec:app-prompts}

\setlength{\fboxsep}{1pt}
\newcommand{\mybox}[2]{\colorbox{#1!20}{#2}}

\begin{tcolorbox}[colback=gray!20, colframe=darkgray, title=\textsc{RePro} Prompt]
Your task is to read and paraphrase the provided text following these instructions:

- Delete clearly irrelevant content:

\quad- Website headers, navigation bars, or menu items (e.g., ``Home | About | Contact'')
  
\quad- Unrelated HTTP links (e.g., ads, trackers, developer tools)
  
\quad- Generic footers (e.g., contact info, privacy policies, unsubscribe links)
  
\quad- Empty lines or decorative elements (e.g., ``---'')
  
- Preserve all content that is relevant and meaningful:

\quad- Informative or independently useful
  
\quad- Related to the topic, even tangentially
  
\quad- Provides context, background, or supporting value
  
\quad- Includes technical terms, key concepts, factual details, reasoning, and examples
  
- Handle mixed-relevance sentences carefully:

\quad- Remove only the irrelevant fragment if the rest remains coherent
  
\quad- Delete the whole sentence if the remainder loses meaning
  
- Do not alter meaningful content unnecessarily:

\quad- Only delete or modify when content is clearly meaningless or off-topic
  
\quad- Preserve the original structure, logic, and depth of the text
  
- Do not add explanations, notes, assumptions, or claims not found in the original text

Here is the text:

\mybox{blue}{\{Organic Text\}}

Task:

After thoroughly reading the above text, paraphrase it in high-quality and clear English following the instructions.

Start your response immediately with ``Here is a paraphrased version:'' and then provide the paraphrased text.
\end{tcolorbox}

\begin{tcolorbox}[colback=gray!20, colframe=darkgray, title=DataMan Prompt]
Please score the text on fourteen evaluation criteria and specify its domain:

Text: \mybox{blue}{\{Text\}}

Domain:\_

[1]Accuracy:\_/5

[2]Coherence:\_/5

[3]Language Consistency:\_/5

[4]Semantic Density:\_/5

[5]Knowledge Novelty:\_/5

[6]Topic Focus:\_/5

[7]Creativity:\_/5

[8]Professionalism:\_/5

[9]Style Consistency:\_/5

[10]Grammatical Diversity:\_/5

[11]Structural Standardization:\_/5

[12]Originality:\_/5

[13]Sensitivity:\_/5

[14]Overall Score:\_/5
\end{tcolorbox}

\begin{tcolorbox}[colback=gray!20, colframe=darkgray, title=Structure Prompt]
[Instruction]

You are given two pieces of text: an original pretraining data sample and a rephrased version.

Your task is to judge if the rephrased version preserves the **structure** of the original sample.

- By ``structure'', we mean formatting, style, and presentation (e.g., paragraphing, JSON, list format, code blocks, markdown usage, plain text style).

- Do NOT consider semantic meaning. Ignore whether the words are the same or the content is equivalent.

- Focus only on whether the rephrased sample follows the same textual structure as the original (e.g., if the original is plain text paragraphs, the rephrased should also be plain text; if the original has bullet lists, the rephrased should also have bullet lists).

[Output]

Output **only** `1` if the structure is preserved.

Output **only** `0` if the structure is not preserved.

[Examples]

Example 1:

Original:

This is a paragraph.

This is another line.

Rephrased:

Here is a rewritten paragraph.

Here is another line of text.

Explanation: Both are plain text paragraphs, no special formatting. Structure preserved. Output: 1

---

Example 2:

Original:

- Item one

- Item two

Rephrased:

First item. Second item.

Explanation: The original uses a bullet list, while the rephrased is plain sentences. Structure not preserved. Output: 0

---

Example 3:

Original:

\{\{``name'': ``Alice'', ``age'': 30\}\}

Rephrased:
\{\{``person'': ``A.'', ``years'': 30\}\}

Explanation: Both are JSON objects with the same structured format. Structure preserved. Output: 1

Explanation: The original is plain code with no markdown fences, while the rephrased introduces code fences. Structure not preserved. Output: 0

[Original]

\mybox{blue}{\{Organic Text\}}

[Rephrased]

\mybox{blue}{\{Recycled Text\}}
\end{tcolorbox}

\begin{tcolorbox}[colback=gray!20, colframe=darkgray, title=Structure Classification Prompt]
Consider the following web page:

Content:

\mybox{blue}{\{Text\}}

Your task is to carefully classify the structure of the given web page. Here, structure refers to the way the content is represented — for example, its markup language, formatting conventions, or encoding style — rather than its topic or purpose. The structure should NOT be HTML.

Please respond with only the name of the structure type, without any additional explanation, commentary, or extra text.
\end{tcolorbox}

\begin{tcolorbox}[colback=gray!20, colframe=darkgray, title=Operation Classification Prompt]
Based on the original text and its rephrased version, extract the key operations that were performed to transform the original text into the rephrased text.

Each operation should be described as one verb + one noun, e.g., ``removing ads''. 

Focus on significant changes such as rewording, restructuring, removing, or clarifying content, while ignoring minor edits like punctuation or spacing adjustments.

Respond strictly in JSON format:
\{

\quad\quad ``operations'': [

\quad\quad\quad\quad operation1,

\quad\quad\quad\quad operation2,

\quad\quad\quad\quad operation3,

\quad\quad\quad\quad ...

\quad\quad ]
  
\}

[Original Text]: \mybox{blue}{\{Organic Text\}}

[Rephrased Text]: \mybox{blue}{\{Recycled Text\}}
\end{tcolorbox}

\begin{tcolorbox}[colback=gray!20, colframe=darkgray, title=WRAP Prompt]
For the following paragraph give me a diverse paraphrase of the same in high quality English language as in sentences on Wikipedia.

Here is the paragraph:

\mybox{blue}{\{Text\}}

Start your response immediately with ``Here is a paraphrased version:'' and then provide the paraphrased text.
\end{tcolorbox}

\begin{tcolorbox}[colback=gray!20, colframe=darkgray, title=ReWire Prompt, label=prompt:wrap]
Below is a draft from an AI Assistant when trying to accomplish task or solving a problem. Analyze and understand the task and problem(s) to be solved. Then pretend to be the expert who is most skillful to acomplish this task, write down the detailed thinking process and internal monologue that went into identifying a strategy and lay out a plan about how to solve this problem. Experts usually apply meta-reasoning and planning to reason about how to best accomplish the task before jumping to solution.

Deliberate meta-reasoning also involves reflection which can help identify issues and take a step back to explore other paths. Below are some generic examples of starting questions experts could ask themselves during meta-reasoning process. The expert will come up with the most relevant questions that can help with their thinking process, which are also very specific to the task.

Let's first try to understand the task and exactly what problem(s) to be solved. What is the core issue or problem that needs to be addressed? What are the key assumptions underlying this problem?

How can I break down this problem into smaller, more manageable parts? How can I simplify the problem so that it is easier to solve?

What kinds of solution typically are produced for this kind of problem specification? Given the problem specification and the current best solution, have a guess about other possible solutions. Let's imagine the current best solution is totally wrong, what other ways are there to think about the problem specification.

What is the best way to modify this current best solution, given what you know about these kinds of problem specification?

Am I on the right track? Let's check our progress so far.

Let's make a step by step plan and implement it with good notion and explanation.

Finally, write an improved response after thinking about how to accomplish the task. Take information and details from the original draft whenever they are useful. Therefore, the improved response should not be shorter than the original response. The improved response should have better formatting and readability, with more coherent and in-depth reasoning, while removing any noise or digression. Note that the best experts chosen to answer each prompt may be different, so please make sure the you do not sound like the same expert for all tasks.

IMPORTANT: Start your analysis and thinking right away. DO NOT add any filler text, explanations or notes about your response. Put the thinking and planning between \textless thinking starts\textgreater~and \textless thinking ends\textgreater, and the improved response between \textless improved response starts\textgreater~and \textless improved response ends\textgreater.

Original Draft: \mybox{blue}{\{Text\}}
\end{tcolorbox}

\begin{tcolorbox}[colback=gray!20, colframe=darkgray, title=\textsc{RePro} Prompt + Reward Descriptions]
Your task is to read and paraphrase the provided text following these instructions:

- Delete clearly irrelevant content:

\quad- Website headers, navigation bars, or menu items (e.g., ``Home | About | Contact'')
  
\quad- Unrelated HTTP links (e.g., ads, trackers, developer tools)
  
\quad- Generic footers (e.g., contact info, privacy policies, unsubscribe links)
  
\quad- Empty lines or decorative elements (e.g., ``---'')
  
- Preserve all content that is relevant and meaningful:

\quad- Informative or independently useful
  
\quad- Related to the topic, even tangentially
  
\quad- Provides context, background, or supporting value
  
\quad- Includes technical terms, key concepts, factual details, reasoning, and examples
  
- Handle mixed-relevance sentences carefully:

\quad- Remove only the irrelevant fragment if the rest remains coherent
  
\quad- Delete the whole sentence if the remainder loses meaning
  
- Do not alter meaningful content unnecessarily:

\quad- Only delete or modify when content is clearly meaningless or off-topic
  
\quad- Preserve the original structure, logic, and depth of the text
  
- Do not add explanations, notes, assumptions, or claims not found in the original text

---------------------------------------

REQUIRED CRITERIA:

**1. Data Quality**  
When paraphrasing, keep in mind several general quality aspects.  
Strive for accuracy so the rewritten content reflects the intended meaning of the original.  
Maintain coherence so the text flows logically and remains easy to follow.  
Use clear, consistent, and professional language that aligns with the tone of the original.  
Aim for focused and precise wording without unnecessary repetition or drift.  
These aspects are meant to guide the paraphrasing process rather than enforce strict rules.

**2. Semantic Faithfulness**  
The paraphrased text should preserve all essential meaning, intent, and information from the original.  
It should not introduce new ideas, remove key points, or reinterpret the content.

**3. Structural Faithfulness**  
The paraphrased output should follow the same structure and formatting style as the original text.  
Paragraph breaks, list formats, and any JSON, code blocks, or markdown elements should be maintained to keep the organizational layout consistent.

**4. Length Faithfulness**  
The paraphrased text should remain close in length to the original and maintain a similar level of detail.  
Do not add new content, explanations, or elaborations, and do not significantly expand or shorten the material.

---------------------------------------

Here is the text:

\mybox{blue}{\{Organic Text\}}

Task:

After thoroughly reading the above text, paraphrase it in high-quality and clear English following all instructions and criteria.

Start your response immediately with ``Here is a paraphrased version:'' and then provide the paraphrased text.
\end{tcolorbox}

\newpage
\section{Case Study}
\label{sec:app-cases}

\subsection{Behavior of \textsc{RePro} Recycling}
\label{sec:case-repro}

In this section, we present more representative examples showing how \textsc{RePro} transforms organic data into higher-quality rephrasings. The quality percentile reflects how an example ranks relative to others in the pool based on its fastText score, with a higher percentile indicating better quality.
For low-quality organic data, we observe that \textsc{RePro} can clarify unclear information or logical gaps. 
For moderate-to-low quality data that often contains irrelevant webpage artifacts (such as navigation text or image placeholders), \textsc{RePro} removes these elements during rephrasing while keeping the essence of the text.
For already high-quality data, \textsc{RePro} reorganizes and paraphrases the content to improve clarity and further raise the quality score. Across these cases, as well as many more examined through manual inspection, we did not observe issues such as degeneration, repeated formatting patterns, or reduced diversity. Through RL optimization, \textsc{RePro} preserves the key characteristics of organic data (writing style, structure, genre, etc.) and learns to flexibly apply various operations to polish the text.

% Set the width of the column separation rule
\setlength{\columnseprule}{0.7pt}
% Set the color of the rule (optional, default is black)
\def\columnseprulecolor{\color{gray!70}}

\begin{tcolorbox}[
    colback=red!5!white,
    colframe=red!75!black,
    title=\textsc{RePro}: Low Quality $\rightarrow$ Moderate Quality\\Operations: \textit{Remove, Paraphrase}
]
\begin{multicols}{2}

\textbf{Organic: Quality Percentile=16.3\%} \\[4pt]
Kate Upton ‘Bombshell Beauty’ Represents Us Elle September 2013! (Pics) \\
Image00001American swim suit model and actress Kate Upton represents Elle September for 2013. She has been photographed by Carter Smith and styled by Joe Zee.Image00001Image00001Image00001 \\
Speak Your Mind

\columnbreak

\textbf{Recycled: Quality Percentile=33.2\%} \\[4pt]
Kate Upton, the renowned American swimwear model and actress, graces the cover of Elle magazine for September 2013. Photographed by Carter Smith, she is styled by Joe Zee, embodying the essence of "Bombshell Beauty" in the issue.

\end{multicols}
\end{tcolorbox}

\begin{tcolorbox}[
    colback=red!5!white,
    colframe=red!75!black,
    title=\textsc{RePro}: Low Quality $\rightarrow$ High Quality\\Operations: \textit{Remove, Clarify}
]
\begin{multicols}{2}

\textbf{Organic: Quality Percentile=4.8\%} \\[4pt]
Interview Question \\
Interview Poway, CA \\
Shift 0xFF to the right by 2, what do you get? \\
Interview Answer \\
1 Answer \\
i THINK since FF is 1111 1111, and you get 0011 1111, you end up with 0x3F \\
Interview Candidate on Nov 10, 2011 \\
Add Answers or Comments \\
To comment on this, Sign In or Sign Up.

\columnbreak

\textbf{Recycled: Quality Percentile=97.9\%} \\[4pt]
Interview Question \\
Interview Location: Poway, CA \\
Question: Shift the value 0xFF to the right by 2 bits. What is the result? \\
Answer: \\
Consider the binary representation of 0xFF, which is 1111 1111. When you shift this binary number to the right by 2 bits, the result is 0011 1111. This binary number corresponds to the hexadecimal value 0x3F. \\
Interview Candidate: November 10, 2011

\end{multicols}
\end{tcolorbox}

\begin{tcolorbox}[
    colback=red!5!white,
    colframe=red!75!black,
    title=\textsc{RePro}: Moderate Quality $\rightarrow$ High Quality\\Operations: \textit{Remove, Paraphrase}
]
\begin{multicols}{2}

\textbf{Organic: Quality Percentile=35.8\%} \\[4pt]
Observation and Growth in Scientific Knowledge \\
In the writings of scientists we find claim to the effect that we can observe items such as pulsars, gravity waves, quarks, electrons, etc. An epistemological theory, originally developed by Dretske and modified by Jackson, is used to give an account of such claims and the extent to which they may be deemed correct. The theory eschews talk of the theory-ladenness of observation while giving an account of how our observation reports may evolve with growth in scientific knowledge. The theory is partially linked to Shapere's recent account of observation in science. \\
Keywords No keywords specified (fix it) \\
Categories (categorize this paper) \\
DOI 10.2307/193125 \\
Save to my reading list \\
Follow the author(s) \\
My bibliography \\
Export citation \\
Find it on Scholar \\
Edit this record \\
Mark as duplicate \\
Revision history Request removal from index \\
Download options

\columnbreak

\textbf{Recycled: Quality Percentile=95.7\%} \\[4pt]
Observation and Growth in Scientific Knowledge \\
In scientific literature, we often encounter assertions about the ability to observe phenomena like pulsars, gravity waves, quarks, and electrons. This discussion explores an epistemological theory originally proposed by Dretske and later refined by Jackson, which is employed to analyze such claims and evaluate their validity. The theory avoids the notion that observation is inherently theory-laden, instead explaining how observation reports can evolve alongside the expansion of scientific knowledge. It also partially connects to Shapere's recent approach to understanding observation in scientific contexts.

\end{multicols}
\end{tcolorbox}

\begin{tcolorbox}[
    colback=red!5!white,
    colframe=red!75!black,
    title=\textsc{RePro}: High Quality $\rightarrow$ Higher Quality\\Operations: \textit{Reorganize, Paraphrase}
]
\begin{multicols}{2}

\textbf{Organic: Quality Percentile=83.2\%} \\[4pt]
Preparation Instructions \\
Preheat oven to 325°F. Heat 2 Tablespoons peanut oil in a large (5 quart+) oven proof Dutch oven over medium heat. Lightly brown meat in batches. Do not over crowd or meat will just release juices and not brown. Transfer browned meat to a large bowl. Put remaining 1Tablespoon of peanut oil into the pot and sauté the onions, garlic and ginger until the onions are clear. Meanwhile, thoroughly whisk together the Cake Spice, ketchup, sherry, tamari or soy sauce, sesame oil and honey. Taste, may taste a bit salty but the red potatoes will absorb this salt. Adjust for honey. Put onion mixture and potatoes into the large bowl. Pour the whisked sauce over the stew and stir or toss to coat. Pour stew back into the Dutch oven and bake in the oven for 3 hours or until the meat and vegetables are tender. To serve garnish with sliced green onions.

\columnbreak

\textbf{Recycled: Quality Percentile=95.1\%} \\[4pt]
Preparation Instructions \\
1. Preheat your oven to 325°F. In a large ovenproof Dutch oven (5 quarts or larger), heat 2 tablespoons of peanut oil over medium heat. Brown the meat in batches without overcrowding, as this will prevent the juices from releasing and ensure proper browning. Transfer the browned meat to a large bowl. \\
2. In the same pot, add the remaining 1 tablespoon of peanut oil and sauté the onions, garlic, and ginger until the onions become transparent. \\ 
3. Meanwhile, whisk together the Cake Spice, ketchup, sherry, tamari, or soy sauce, sesame oil, and honey. The mixture may taste slightly salty, but the red potatoes will absorb the excess. Adjust the seasoning with honey to your preference. \\ 
4. Combine the sautéed onion mixture with the potatoes in the large bowl and pour the whisked sauce over the stew, stirring or tossing to ensure everything is well-coated. \\
5. Transfer the stew back into the Dutch oven and bake for 3 hours, or until the meat and vegetables are tender. Serve garnished with sliced green onions.

\end{multicols}
\end{tcolorbox}

% An interesting case study in Appendix~\ref{sec:app-cases} further illustrates this phenomenon,

\subsection{Behavior of Different Rephrasing Methods}
\label{sec:case-all}

Finally, we study the behavior of different rephrasing methods.
The recycled text from ReWire or the rephraser trained without faithfulness rewards introduces extraneous information loosely related to the original text (e.g., the assassination of Archduke Franz Ferdinand) and alters the structure of the organic data from markdown to plain text. This may increase the quality score but reduce the overall faithfulness of recycled data, posing a risk of model collapse.
WRAP also adds content that is not present in the organic data.
In contrast, \textsc{RePro} strictly adheres to the original content and structure while improving the writing quality by removing irrelevant advertisements and metadata.

\begin{tcolorbox}[colback=cyan!20, colframe=blue, title=Organic Data]
Related Ads\\\\5 Facts About Gallipoli\\\\Here are 5 facts about Gallipoli\\\\  - Gallipoli is a peninsular in Turkey\\  - British and colonial forces attacked the peninsula in 1915 during World War I\\  - It was known as the Dardanelles Campaign\\  - Deaths were 43,000 British, 15,000 French, 8,700 Australians, 2,700 New Zealanders and 1,370 Indians\\  - The Allies withdrew from the campaign which is celebrated on ANZAC Day\\blog comments powered by Disqus","url":"http://5factsabout.com/articles/5-facts-about-Gallipoli/","warcinfo":"robots: classic\\hostname: ip-10-137-6-227.ec2.internal\\software: Nutch 1.6 (CC)/CC WarcExport 1.0\\isPartOf: CC-MAIN-2015-40\\operator: CommonCrawl Admin\\description: Wide crawl of the web for September 2015\\publisher: CommonCrawl\\format: WARC File Format 1.0\\conformsTo: http://bibnum.bnf.fr/WARC/WARC\_ISO\_28500\_version1\_latestdraft.pdf
\end{tcolorbox}

\begin{tcolorbox}[colback=cyan!20, colframe=blue, title=Recycled Data (\textsc{RePro})]
**Gallipoli: Key Historical Facts**

Gallipoli is a peninsula in Turkey that played a significant role in military history. 

During World War I, British and colonial forces launched an attack on the region in 1915 as part of the Dardanelles Campaign. 

The campaign resulted in substantial casualties, including 43,000 British, 15,000 French, 8,700 Australian, 2,700 New Zealand, and 1,370 Indian soldiers. 

Despite initial advances, the Allies ultimately withdrew from the campaign. 

The event is commemorated annually on ANZAC Day.
\end{tcolorbox}

\begin{tcolorbox}[colback=cyan!20, colframe=blue, title=Recycled Data (\textsc{RePro} w/ SFT)]
5 Key Points About Gallipoli\\\\1. **Location**: Gallipoli is a peninsula situated in Turkey.\\2. **Historical Conflict**: It was the site of an attack by British and colonial forces during World War I in 1915, known as the Dardanelles Campaign.\\3. **Casualties**: The campaign resulted in significant losses, with 43,000 British troops, 15,000 French soldiers, 8,700 Australians, 2,700 New Zealanders, and 1,370 Indian soldiers killed.\\4. **Military Withdrawal**: Due to the heavy casualties and strategic challenges, the Allies decided to withdraw from the campaign.\\5. **Memorial Day**: This historical event is commemorated annually on ANZAC Day.
\end{tcolorbox}

\begin{tcolorbox}[colback=cyan!20, colframe=blue, title=Recycled Data (Prompting)]
5 Facts About Gallipoli\\\\The following are five facts regarding Gallipoli:\\\\- Gallipoli is a peninsula located in Turkey.\\- British and colonial troops launched an attack on the peninsula in 1915 during World War I.\\- The campaign is referred to as the Dardanelles Campaign.\\- The total number of casualties included 43,000 British, 15,000 French, 8,700 Australians, 2,700 New Zealanders, and 1,370 Indians.\\- The Allied forces eventually withdrew from the campaign, which is commemorated on ANZAC Day.
\end{tcolorbox}

\begin{tcolorbox}[colback=cyan!20, colframe=blue, title=Recycled Data (Prompting w/ Thinking)]
$<$think$>$\\Okay, the user wants me to paraphrase the given text following specific instructions. Let me start by understanding the original text.\\The original text has a heading \"Related Ads\" and then \"5 Facts About Gallipoli\". The main content lists five facts about Gallipoli. There's also a mention of blog comments powered by Disqus, which is a footer or unrelated link. \\First, I need to delete irrelevant content. The \"Related Ads\" header is probably not needed. The \"blog comments powered by Disqus\" is a footer, so I should remove that. Also, check for any empty lines or decorative elements. The text seems to have a bullet point list, which is relevant.\\Now, the task is to preserve all meaningful content. The five facts are important. Let me check each point:\\1. Gallipoli is a peninsular in Turkey \u2013 that's a fact.\\2. British and colonial forces attacked in 1915 during WWI \u2013 relevant.\\3. Known as Dardanelles Campaign \u2013 correct.\\4. Deaths listed \u2013 important details.\\5. Allies withdrew and it's celebrated on ANZAC Day \u2013 key info.\\I need to paraphrase these points without changing the meaning. Also, ensure that the structure is maintained but in a different way. For example, instead of using \"Here are 5 facts\", maybe start with \"Gallipoli, a peninsula in Turkey, was the site of a significant military campaign during World War I.\" Then list the facts concisely.\\Check for any mixed-relevance sentences. The original text has the \"blog comments\" part, which is irrelevant, so remove that. The rest is all relevant. \\Make sure not to add any explanations or assumptions. Just rephrase the existing content. Also, preserve technical terms like \"Dardanelles Campaign\" and the numbers. \\So the final paraphrase should start with the header \"Here is a paraphrased version:\" followed by the rephrased facts. Let me put that together now.\\$<$/think$>$\\Here is a paraphrased version:  \\Gallipoli, a peninsula in Turkey, was the site of a significant military campaign during World War I. British and colonial forces launched an attack in 1915, known as the Dardanelles Campaign. The conflict resulted in approximately 43,000 British, 15,000 French, 8,700 Australian, 2,700 New Zealand, and 1,370 Indian casualties. The Allied forces eventually withdrew from the campaign, which is commemorated on ANZAC Day.
\end{tcolorbox}

\begin{tcolorbox}[colback=cyan!20, colframe=blue, title=Recycled Data (RL w/o Faithfulness)]
Gallipoli is a strategically significant peninsula located in northwest Turkey, situated between the Aegean Sea and the Dardanelles Strait. This location played a pivotal role during the First World War, specifically during the 1915 campaign known as the Dardanelles or the Battle of Gallipoli, a major Allied military operation initiated by Britain and its colonial allies. The objective was to open a new front in the Ottoman Empire to relieve pressure on the Russian Empire and to liberate Constantinople from Turkish control.\\The Gallipoli campaign saw a significant deployment of British, French, Australian, New Zealand, and Indian troops. The operation commenced on May 19, 1915, with the aim of capturing the Turkish-held Gallipoli Peninsula and forcing the Ottomans to withdraw from the Eastern Front. However, the campaign faced numerous challenges, including treacherous terrain, strong Turkish defenses, and logistical difficulties, which contributed to a costly stalemate. \\The Allied forces suffered heavy casualties during the campaign, with estimates ranging from 43,000 British, 15,000 French, 8,700 Australians, 2,700 New Zealanders, and 1,370 Indian soldiers killed, in addition to numerous wounded. Despite their efforts, the Allies were unable to achieve their strategic objectives, and the campaign ended in disaster, leading to the withdrawal of British and French forces from the peninsula in January 1916. \\This pivotal event in military history has been immortalized in Australian and New Zealand military traditions, and ANZAC Day, observed on April 25th, commemorates the service and sacrifice of these nations' soldiers in the First World War, including the gallantry displayed at Gallipoli. The commemorations emphasize themes of bravery, sacrifice, and national identity, serving as a reminder of the significant contributions made by Australian and New Zealand soldiers to the Allied war efforts.
\end{tcolorbox}

\begin{tcolorbox}[colback=cyan!20, colframe=blue, title=Recycled Data (fastText as Quality Reward)]
The text outlines five significant facts regarding Gallipoli. First, it describes Gallipoli as a geographical region located in Turkey. Additionally, it highlights the historical context of the area, specifically the military operations that took place there during World War I. Furthermore, the text provides a detailed account of the casualties sustained during the conflict, emphasizing the substantial loss of life on both sides. Finally, it mentions the historical significance of the campaign, including the eventual withdrawal of the Allied forces and its connection to ANZAC Day. Overall, the text serves as a concise summary of the key points related to Gallipoli's historical importance.
\end{tcolorbox}

\begin{tcolorbox}[colback=cyan!20, colframe=blue, title=Recycled Data (WRAP)]
Five Noteworthy Factoids Concerning Gallipoli\\Gallipoli, a coastal region in Turkey, has been the focus of five intriguing facts that have garnered significant attention.\\Firstly, Gallipoli is a peninsula located in Turkey, and it has become the focal point of historical events, particularly during World War I.\\Secondly, the British and colonial forces launched an attack on the peninsula in 1915, which was known as the Dardanelles Campaign.\\Thirdly, the casualties during the campaign were significant, with 43,000 British, 15,000 French, 8,700 Australians, 2,700 New Zealanders, and 1,370 Indians losing their lives.\\Fourthly, the Allies withdrew from the campaign, marking a significant moment in the history of World War I.\\Finally, the campaign has become a symbol of bravery and sacrifice, and it is celebrated annually on ANZAC Day, which stands for Australian and New Zealand Army Corps Day.
\end{tcolorbox}

\begin{tcolorbox}[colback=cyan!20, colframe=blue, title=Recycled Data (ReWire)]
The Gallipoli campaign was a pivotal event in World War I, marking a significant turning point in the conflict. Located on the Gallipoli peninsula in Turkey, the campaign was fought between the Allied Powers, including British, French, Australian, New Zealand, and Indian forces, and the Ottoman Empire.\\\\To understand the context of the campaign, it is essential to consider the events leading up to the war. In 1914, the assassination of Archduke Franz Ferdinand sparked a chain reaction of diplomatic crises and military mobilizations, eventually leading to the outbreak of World War I. The Allied Powers, seeking to gain a strategic advantage, set their sights on the Dardanelles, a narrow strait that connected the Aegean Sea to the Sea of Marmara, and ultimately, to the Black Sea.\\\\The Gallipoli campaign, also known as the Dardanelles Campaign, began on February 19, 1915, when British and French warships launched a bombardment of the Ottoman defenses. The campaign was intended to capture the strategic location of Constantinople, now Istanbul, and secure a vital trade route. However, the Ottoman forces, led by Mustafa Kemal, were better prepared than anticipated, and the campaign quickly turned into a disastrous defeat for the Allies.\\\\The campaign itself was marked by a series of bloody battles, including the landing at Anzac Cove, the Battle of Lone Pine, and the Battle of Chunuk Bair. Despite being outnumbered, the Ottoman forces held their ground, and the Allies suffered heavy casualties. The death toll was staggering, with 43,000 British, 15,000 French, 8,700 Australian, 2,700 New Zealand, and 1,370 Indian soldiers losing their lives.\\\\The aftermath of the campaign was marked by the withdrawal of the Allied forces in January 1916. The failure of the campaign had significant consequences, including the resignation of Winston Churchill, who had been a strong advocate for the campaign. The legacy of the Gallipoli campaign is still commemorated today, with ANZAC Day, April 25, marking the anniversary of the landing at Anzac Cove.\\\\In conclusion, the Gallipoli campaign was a pivotal event in World War I, marking a significant turning point in the conflict. The campaign's failure had far-reaching consequences, including the resignation of key leaders and a shift in the balance of power. Today, the campaign is remembered as a testament to the bravery and sacrifice of the soldiers who fought and died on the Gallipoli peninsula.
\end{tcolorbox}

% You can have as much text here as you want. The main body must be at most $8$
% pages long. For the final version, one more page can be added. If you want, you
% can use an appendix like this one.

% The $\mathtt{\backslash onecolumn}$ command above can be kept in place if you
% prefer a one-column appendix, or can be removed if you prefer a two-column
% appendix.  Apart from this possible change, the style (font size, spacing,
% margins, page numbering, etc.) should be kept the same as the main body.
%%%%%%%%%%%%%%%%%%%%%%%%%%%%%%%%%%%%%%%%%%%%%%%%%%%%%%%%%%%%%%%%%%%%%%%%%%%%%%%
%%%%%%%%%%%%%%%%%%%%%%%%%%%%%%%%%%%%%%%%%%%%%%%%%%%%%%%%%%%%%%%%%%%%%%%%%%%%%%%

\end{document}